\documentclass{article}

\usepackage[preprint,nonatbib]{style_test}

\usepackage[utf8]{inputenc} 
\usepackage[T1]{fontenc}    
\usepackage{url}            
\usepackage{booktabs}       
\usepackage{amsfonts}       
\usepackage{nicefrac}       
\usepackage{microtype}      
\usepackage{enumitem,kantlipsum}
\usepackage{nameref}
\usepackage{graphicx}
\usepackage{caption} 
\usepackage{subcaption}

\usepackage{xcolor}         
\usepackage{hyperref}       
\usepackage{float}

\usepackage{pifont}
\newcommand{\xmark}{\ding{55}}%

\usepackage{tabularx}

\usepackage[style=nature,backend=biber]{biblatex}
\usepackage{lineno}

\newcommand*\samethanks[2][\value{footnote}]{\footnotemark[2]}

\makeatletter
\def\@fnsymbol#1{\ensuremath{\ifcase#1\or 1 \or 2 \or 3 \or 4 \or \dagger\dagger
   \or \ddagger\ddagger \else\@ctrerr\fi}}
\makeatother
   
\title{Phy-Q as a Measure for Physical Reasoning Intelligence}

\author{%
Cheng Xue$^{1,2}$ \space \space \space
Vimukthini Pinto$^{1,2}$ \space \space \space  Chathura Gamage$^{1,2}$\\ \textbf{Ekaterina Nikonova}$^{1}$ \space \space \space \textbf{Peng Zhang}$^{1}$\space \space \space \textbf{Jochen Renz}$^{1}$ \\}

\begin{document}

\maketitle

\def\thefootnote{1}\footnotetext[1]{ School of Computing, The Australian National University, Canberra, Australia}
\def\thefootnote{2}\footnotetext[2]{ These authors contributed equally}

\begin{abstract}
Humans are well-versed in reasoning about the behaviors of physical objects and choosing actions accordingly to accomplish tasks, while it remains a major challenge for AI. To facilitate research addressing this problem, we propose a new testbed that requires an agent to reason about physical scenarios and take an action appropriately. Inspired by the physical knowledge acquired in infancy and the capabilities required for robots to operate in real-world environments, we identify 15 essential physical scenarios. 
We create a wide variety of distinct task templates, and we ensure all the task templates within the same scenario can be solved by using one specific strategic physical rule. By having such a design, we evaluate two distinct levels of generalization, namely the \textit{local generalization} and the \textit{broad generalization}. We conduct an extensive evaluation with human players, learning agents with varying input types and architectures, and heuristic agents with different strategies. 
Inspired by how human IQ is calculated, we define the physical reasoning quotient (Phy-Q score) that reflects the physical reasoning intelligence of an agent using the physical scenarios we considered.
Our evaluation shows that 1) all agents are far below human performance, and 2) learning agents, even with good local generalization ability, struggle to learn the underlying physical reasoning rules and fail to generalize broadly. 
We encourage the development of intelligent agents that can reach the human level Phy-Q score.
\end{abstract}

\section{Main}

\subsection{Introduction}

The ability to reason about objects’ properties and behaviours in physical environments lies at the core of human cognitive development \cite{Davis2006}. A few days after birth, infants understand object solidity \cite{Valenza2006} and within the first year of birth, they understand notions such as object permanence \cite{permanence}, spatiotemporal continuity \cite{continuity}, stability \cite{stability}, support \cite{support}, causality \cite{causality}, and shape constancy \cite{shapeconstancy}. Generalization performance on novel physical puzzles is commonly used as a measure of physical reasoning abilities for children \cite{Carmel2000,Cheke2012}, animals \cite{Crows1}, and AI agents \cite{Phyre,OGRE,Allen2020,ahmed2020causalworld}.

Chollet's study \cite{Chollet19} on the measure of intelligence proposes a qualitative spectrum of different forms of generalization that includes \textit{local generalization} and \textit{broad generalization}. Current evidence \cite{lg1,lg2,lg3} suggests that contemporary deep learning models are local-generalization systems, i.e., systems that adapt to known unknowns within a single task. \textit{Broad generalization}, on the other hand, can be characterized as `adaptation to unknown unknowns across a broad category of related tasks' and is being increasingly emphasized among the AI research community \cite{Chollet19,geirhos2020shortcut}. Moreover, when solving physics puzzles, it is common that a player must use a strategy to work out a plan and must use dexterity to accurately execute the strategic plan \cite{strategyAndDexterity}. For instance, in a snooker game, a player needs to plan the path of the white cue ball, e.g., where it should go and where to stop, and then execute the strike that precisely produces the planned path. 
One of the views of cognitive psychology researchers is that humans possess inaccurate forward physics prediction models \cite{mccloskey_naive_1983,smith_sources_2013} and hence require practice to improve dexterity, high dexterity requirements of physics tasks make it unfair to compare AI agents' physical reasoning ability with average humans'. For example, when a human player fails a physics puzzle, it is hard to tell if it is due to incorrect physical reasoning or due to the inability of making precise actions. Despite the recent advancement in physical reasoning benchmarks and testbeds \cite{Phyre,OGRE,Allen2020,IntPhys2019,CLEVRER,Cophy,ahmed2020causalworld,Physion}, there is a lack of a benchmark or a testbed with human comparable strategic physics puzzles and one that explicitly evaluates learning agents’ local and broad generalization. 

To close these gaps, we propose a new testbed Phy-Q and the associated Phy-Q score that measures physical reasoning intelligence using the physical scenarios we identified. Inspired by the physical knowledge acquired in infancy and the abilities required by the robots to operate in the real world, we have created a wide variety of tasks with low dexterity requirements in the video game Angry Birds.
We believe the contributions from this paper pave the way for the development of agents with human level strategic physical reasoning capabilities. 

Our main contributions can be summarized as follows:  

\begin{itemize}[leftmargin=*]
    \item \textbf{Phy-Q: A testbed for physical reasoning: } We designed a variety of task templates in Angry Birds with 15 physical scenarios, where all the task templates of a scenario can be solved by following a common strategic physical rule. Then we generated task instances from the templates using a task variation generator. This design allows us to evaluate both the local and the broad generalization ability of an agent. 
    We also define the Phy-Q score, a quantitative measure that reflects physical reasoning intelligence using the physical scenarios we considered.

    \item \textbf{An agent-friendly framework: } We propose a framework, which allows training multi-agent instances simultaneously with accelerated game-play speed up to 50 times.
    
    \item \textbf{Establishing results for baseline agents: } The evaluation consists of nine baseline agents: four of our best performing learning agents, four heuristic-based agents, and a random agent.
    For each of the baseline agents, we present the Phy-Q score, the \textit{broad generalization} performance, and the \textit{local generalization} performance.
     We have collected human player data so that agent performance can be directly compared to human performance. 
    
    \item \textbf{A guidance for agents in AIBIRDS competition: }
    Angry Birds is a popular physical reasoning domain among AI researchers with the long running AIBIRDS competition since 2012 \cite{Renzet2019}. In 2016, Angry Birds was considered to be the next milestone in AI where AI will surpass humans \cite{Grace2018}. A time horizon of four years was predicted and so far such a breakthrough seems very unlikely. 
    In the AIBIRDS competition, heuristic methods generally perform better than their deep learning counterparts, but it remains unclear what has contributed to the gap in the performances. It has also not yet been analysed why current AI agents fall short when compared to humans. By the systematic analysis of agents from the AIBIRDS competition, we show how they need to be improved to achieve human-level performance.
\end{itemize}

\subsection{Background and Related Work} \label{Background and Related Work}

\begin{table}[h!]
  \scriptsize
  \caption{Comparison of Phy-Q with related physics benchmarks and competitions}
  \label{benchmarks-table}
  \centering
  \begin{tabular}{lllllll}
    \toprule
     Test & generalization & categorization & procedurally & destructible & observe outcome & human \\
     Environment & to individual & of tasks to & generated & objects & of a desired & player\\
     & physical scenario/s & physical scenarios & tasks/variations &  & physical action & data\\
    \midrule
    PHYRE \cite{Phyre} & \xmark & \xmark & \checkmark & \xmark & \checkmark & \xmark\\
    Virtual Tools \cite{Allen2020} & \xmark & \checkmark & \xmark & \xmark & \checkmark & \checkmark \\
    OGRE \cite{OGRE} & \xmark & \xmark & \checkmark & \xmark & \checkmark & \xmark\\
    \midrule
    IntPhys 2019 \cite{IntPhys2019} & \checkmark & \checkmark & \checkmark & \xmark & \xmark & \checkmark\\
    CLEVERER \cite{CLEVRER} & \checkmark & \xmark & \checkmark & \xmark & \xmark & \xmark \\
    CATER \cite{CATER} & \checkmark & \checkmark & \checkmark & \xmark & \xmark & \xmark\\
    Physion \cite{Physion} & \checkmark & \checkmark & \checkmark & \xmark & \xmark & \checkmark\\
    COPHY \cite{Cophy} & \checkmark & \checkmark & \checkmark & \xmark & \xmark & \checkmark\\
    \midrule
    CausalWorld \cite{ahmed2020causalworld} & \checkmark & \checkmark & \checkmark & \xmark & \checkmark & \xmark\\
    RLBench \cite{james2020rlbench} & \xmark & \xmark & \checkmark & \xmark & \checkmark & \xmark\\
    \midrule
    Computational Pool \cite{Computational_Pool}& \xmark & \xmark & \xmark & \xmark & \checkmark & \xmark\\
    Geometry Friends \cite{GeometryFriends} & \xmark & \xmark & \checkmark & \xmark & \checkmark & \xmark\\
    \midrule
    AIBIRDS \cite{AIBirds}& \xmark & \xmark & \checkmark & \checkmark & \checkmark & \checkmark\\
    Phy-Q (ours) & \checkmark & \checkmark & \checkmark & \checkmark & \checkmark & \checkmark\\
    \bottomrule
  \end{tabular}
\end{table}

In this section, we conduct a comparison between ten related physical reasoning benchmarks and two physics-based AI game competitions to show how Phy-Q testbed advances upon existing work. The comparison is done with respect to six criteria.
1) Measuring \textit{broad generalization} in individual physical scenario/s, i.e, testing the ability of an agent in generalizing to tasks that require the same physical rule to solve.
2) Categorization of tasks of the test environment into different physical scenarios, i.e, agents can be evaluated for individual scenarios to recognize the scenarios that they can perform well.
3) Procedural generation of tasks or variations of the tasks, i.e., the tasks/variants of the tasks in the test environment are created algorithmically facilitating the users to generate any amount of data.
4) Destructibility of objects in the environment, i.e., if the environment contains objects that can be destroyed upon the application of forces. Having destructible objects makes the environment more realistic than an environment that only has indestructible objects since the agents need to consider the magnitude of the force that is applied to the objects. For example, when a robot moves a cup, it needs to reason that the force to exert should be large enough to grab the cup but not too large to break it.
5) Observing the outcome of a desired physical action, i.e., if an agent can physically interact and observe the outcome of the action the agent takes. 
6) Inclusion of human player data, i.e., if the evaluation has results of human players. 

We consider PHYRE \cite{Phyre}, Virtual Tools game \cite{Allen2020}, and OGRE \cite{OGRE}, which are game based benchmarks, IntPhys \cite{IntPhys2019}, CLEVERER \cite{CLEVRER}, CATER \cite{CATER}, and Physion \cite{Physion} which are video based benchmarks, COPHY \cite{Cophy} which is an image based benchmark, CausalWorld \cite{ahmed2020causalworld} and RLBench \cite{james2020rlbench} which are robotic benchmarks. The AI game competitions we consider are Computational Pool \cite{Computational_Pool}, and Geometry Friends \cite{GeometryFriends}. We also included the AIBIRDS \cite{AIBirds} competition for the comparison to show what properties in Phy-Q facilitate the systematic evaluation of AIBIRDS competition agents. Table \ref{benchmarks-table} summarises the comparison.

The most closely related physical reasoning test environment to ours is PHYRE \cite{Phyre}, which also consists of tasks to measure two levels of generalization of agents. PHYRE benchmark tests if agents can generalize to solve tasks within a task template (within template) and if agents can generalize between different task templates (cross template).
The cross template evaluation in PHYRE does not guarantee that the physical rules required to solve the testing tasks exist in the training tasks. This leads to uncertainties in understanding agents' performance: inferior performance may not be an indicator of inferior physical reasoning but of a difficult training and testing split. Whereas the \textit{broad generalization} evaluation in our testbed always ensures that physical rules required in testing tasks are covered in the training tasks, hence we guarantee a more systematic evaluation of the physical reasoning capabilities of AI agents. 
According to the task design in PHYRE, tasks need to be solved by trial and error. That is, even when the physical rule is known, multiple attempts are still needed to solve the tasks. Therefore, PHYRE promotes developing agents with physical dexterity. In contrast, we focus on strategy based physical reasoning tasks that can be solved by a single attempt when the physical rule is understood. We promote developing agents that can understand a physical rule rather than taking a precise action in a physical environment (i.e., agents with strategic physical reasoning capabilities).
Furthermore, a limited number of object shapes, motion, and material properties on scene dynamics hinder the ability of a comprehensive evaluation, as performing well on these tests might not indicate greater physical reasoning ability in more general and realistic contexts \cite{Physion}. Therefore, compared to PHYRE,  in Phy-Q we have 1) 3 more object shapes (rectangles, squares, and triangles) to allow more diverse physical dynamics, 2) destructible objects to make our environment more realistic, and 3) objects with three different materials that have different densities, bounciness, and friction to allow physical reasoning in a more realistic context.

As a recent visual and physical prediction benchmark, Physion \cite{Physion} evaluates algorithms' physical prediction capability using videos of eight different physical scenarios. Compared to Physion, Phy-Q testbed has a more comprehensive set of 15 physical scenarios enabling the evaluation of agents in a wider range of physical scenarios. Phy-Q testbed requires agents to interact with the environment and select the desired action to accomplish physical tasks. Therefore, on top of predicting a physical event's outcome, agents need to apply the acquired physical knowledge to solve new situations, which is considered to be a more advanced type of task in Bloom’s taxonomy \cite{bt2001}. In addition, a study on forward prediction for physical reasoning \cite{Rohit2020} confirms that higher forward prediction accuracy does not necessarily increase performance in domains that require selecting an action. Therefore, the research problem focused on Physion differs from that of Phy-Q.

Despite Angry Birds being a simplified and controlled physics environment as compared to the much messier real physical world, no AI system has been developed that comes close to human performance. To encourage the development of AI agents that can reason with physics as humans do, the AIBIRDS competition has been organized annually since 2012, mostly held at the International Joint Conference on Artificial Intelligence \cite{AIBirds}. Since then, many different AI approaches have been proposed, ranging from modern deep reinforcement learning methods to more old-school heuristic methods, e.g. qualitative physical reasoning methods. 
However, none of these approaches has reached the milestone of achieving human level performance. One major reason is that an agent's performance in the competition does not enable an agent developer to identify the physical scenarios that the agent falls short of. This is because the tasks in the competition are generally complex with multiple physical scenarios within the same task. In this work, we show how the Phy-Q testbed can be used towards guiding the competition agents through a systematic evaluation of agents' performance.

\subsection{Phy-Q Testbed} \label{Section Hi-Phy Physical Reasoning Benchmark}
In this section, we introduce our testbed and discuss the physical scenarios we have identified.  

\subsubsection{Introduction to the Phy-Q Testbed}

Based on the 15 identified physical scenarios (discussed in detail in Section \nameref{Section Physical scenarios in Phy-Q}), we develop a physical reasoning testbed using Angry Birds. In Angry Birds, the player interacts with the game by shooting birds at the pigs from a slingshot. The goal of the player is to destroy all the pigs using the provided set of birds. As the original game by Rovio Entertainment is not open-sourced, we use a research clone of the game developed in Unity \cite{Ferreira2014}. 
The game environment is a deterministic 2D world where objects in motion follow Newtonian physics. 
The game objects are of four types: birds, pigs, blocks, and platforms. There are five types of birds, four of which have powers that can be activated once tapped in their flight. There are three types of pigs varying in size, the health points of the pigs increase with the increase in size. Blocks in the game are made of three materials (wood, ice, and stone) and each of them has 12 variations in shape. Platforms are static objects that remain at a fixed position and are not affected by forces and are indestructible. All other objects are dynamic, i.e., they can be moved by applying forces. Dynamic objects have health points that get reduced upon collisions with other objects and they get destroyed and disappear when health points reach zero. The initial state of a game level is physically stable (i.e., none of the objects is in motion) and the goal is not achieved. The action of an agent is to shoot the bird on the slingshot by providing the release coordinates relative to the slingshot. We have included a module that aids trajectory planning to reduce the dexterity requirement. Additionally, the agent provides the tap time of the bird to activate powers (if available). The selection of release point and tap time makes the action space essentially continuous. When playing, an agent takes a sequence of actions, i.e., shoots the birds in a predefined order. The agent passes a game level when it destroys all pigs with the provided set of birds and fails otherwise. We do not provide the full world state that includes the exact location of objects in the simulator or their physical properties such as mass and friction to the agents as these properties are not directly observable in the real world. Instead, an agent can request screenshots and/or a symbolic representation of the game level at any time while playing. A game screenshot is a 480 x 640 coloured image and the symbolic representation is in JSON format containing all objects in the screenshot represented as a polygon of its vertices (provided in order) and its respective colour map. The colour map provides the list of 8-bit quantized colours that appear in the game object with their respective percentages.

\subsubsection{Physical Scenarios in Phy-Q Testbed} \label{Section Physical scenarios in Phy-Q}

\begin{figure}[t]
  \centering
  \includegraphics[width=0.98\linewidth]{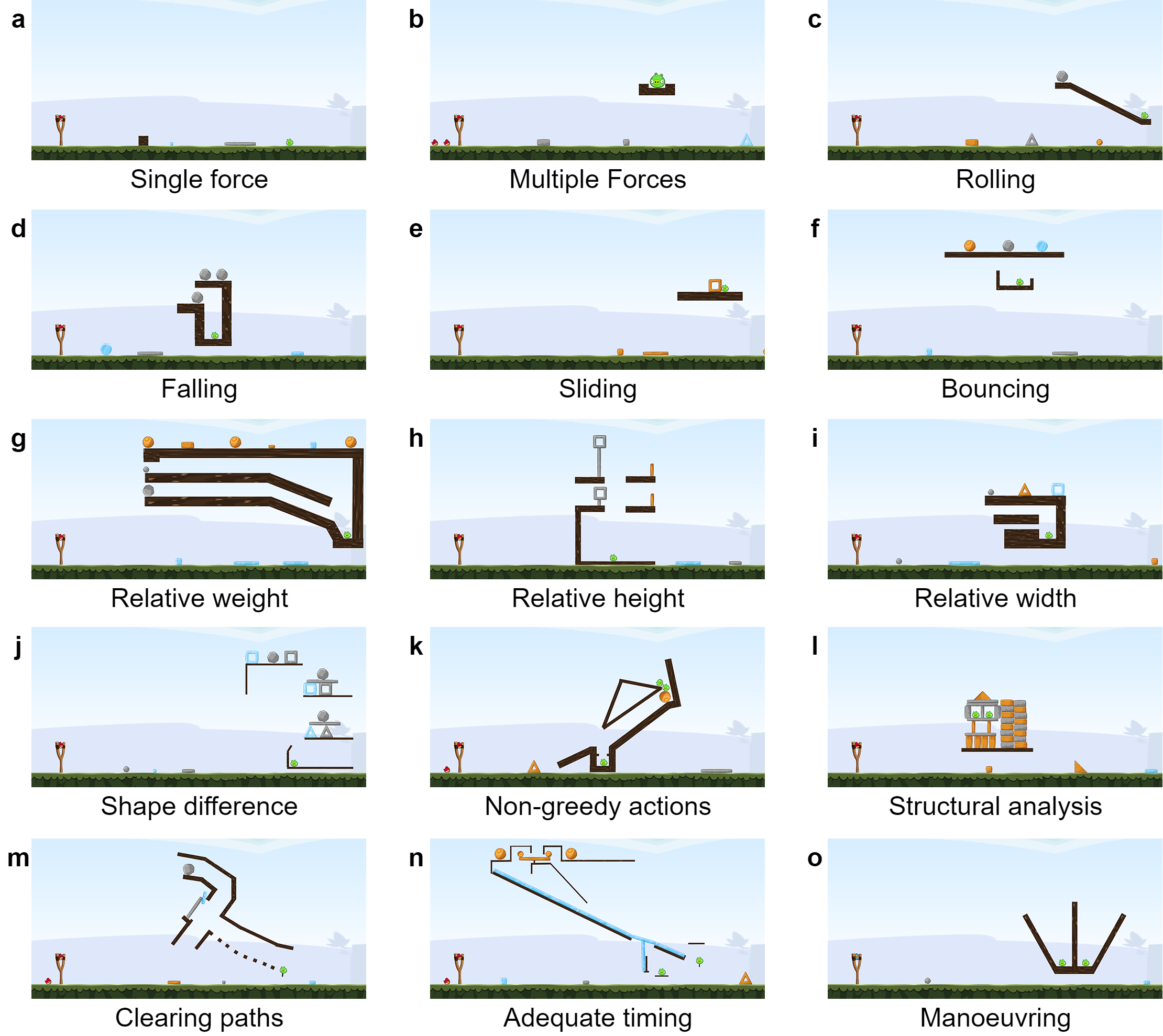}
  \caption{15 example tasks in Phy-Q representing the 15 physical scenarios. The slingshot with birds is situated on the left of the task. The goal of the agent is to kill all the green-coloured pigs by shooting birds from the slingshot. The dark-brown objects are static platforms. The objects with other colours are dynamic and subject to the physics in the environments.}
  \label{fig_six_example_tasks_in_hiphy}
\end{figure}

In this section, we introduce the 15 physical scenarios we consider in our testbed. 
Firstly, we consider the basic physical scenarios associated with applying forces directly on the target objects, i.e., the effect of a single force and the effect of multiple forces \cite{force}. On top of the application of single force, we also include scenarios associated with more complex motion including rolling, falling, sliding, and bouncing, which are inspired by the physical reasoning capabilities developed in human infancy \cite{rolling_falling}. Furthermore, we define the objects' relative weight \cite{wang_williamson_meltzoff_2018}, the relative height \cite{baillargeon_devos_1991}, the relative width \cite{wang_2004}, the shape differences \cite{baillargeon_li_ng_yuan_2008}, and the stability \cite{wilcox_chapa_2004} scenarios, which require physical reasoning abilities infants acquire typically in a later stage.
On the other hand, we also incorporate clearing path, adequate timing, and manoeuvring \cite{Robotic_Challenges}, and taking non-greedy actions \cite{non_greedy}, which are required to overcome challenges for robots to work safely and efficiently in physical environments. 
Each of these scenarios tests a different aspect of the agent's skill, physical understanding, and planning ability. To sum up, the physical scenarios we consider and the corresponding high-level strategic physical rules that can use to achieve the goal of the associated tasks are mentioned below. Example task templates from those scenarios are shown in Figure 1.

\begin{enumerate} 
\item \textbf{Single force: } Target objects have to be destroyed with a single force.
\item \textbf{Multiple forces: } Target objects need multiple forces to destroy.
\item \textbf{Rolling: }Circular objects have to be rolled along a surface to a target. 
\item \textbf{Falling: }Objects have to fall onto a target.
\item \textbf{Sliding: }Non-circular objects have to be slid along a surface to a target.
\item \textbf{Bouncing: }Objects have to be bounced off a surface to reach a target. 
\item \textbf{Relative weight: }Objects with the correct weight have to be moved to reach a target.
\item \textbf{Relative height: }Objects with the correct height have to be moved to reach a target.
\item \textbf{Relative width: }Objects with the correct width or the opening with the correct width have to be selected to reach a target.
\item \textbf{Shape difference: }Objects with the correct shape have to be moved/destroyed to reach a target. 
\item \textbf{Non-greedy actions: }Actions have to be selected in the correct order based on physical consequences. The immediate action may be less effective in the short term but advantageous in long term. i.e., reach fewer targets in the short term to reach more targets later. 
\item \textbf{Structural analysis: }The correct target has to be chosen to break the stability of a structure. 
\item \textbf{Clearing paths: }A path have to be created before the target can be reached.
\item \textbf{Adequate timing: }Correct actions have to be performed within time constraints.
\item \textbf{Manoeuvring: }Objects have to be carefully guided to reach a target.
\end{enumerate}

\subsection{Conclusion and Future Work}

The goal of the Phy-Q testbed is to facilitate the development of physical reasoning AI methods with broad generalizing abilities similar to that of humans. 
As mentioned previously, humans may possess inaccurate forward physics prediction models. We focus on tasks that can be solved by using a strategic physical rule and with low dexterity requirements instead of tasks that require precise forward prediction.
Therefore, towards that goal, we designed 75 task templates considering 15 different physical scenarios in our testbed.
The tasks that belong to the same physical scenario can be solved by a specific strategic physical rule, enabling us to measure the \textit{broad generalization} of agents by allowing the agent to learn a strategic physical rule in the learning phase that can be used in the testing phase. Apart from the \textit{broad generalization} performance evaluation, the Phy-Q testbed also enables evaluating agents' \textit{local generalization} performance.
We have established baseline results from the testbed and have shown that even though current learning agents can generalize locally, the \textit{broad generalization} ability of these agents is below heuristic agents and is far below human performance. 
Further, we have defined Phy-Q score, a score to reflect the physical reasoning ability of agents.
In addition, we have shown how the testbed can be used for the advancement of the AIBIRDS competition agents.    

Although we discourage developing heuristic agents with hard-coded rules that apply only to Angry Birds, we believe the superior performance of these rule-based systems, given that none of the agent developers has seen the Phy-Q tasks previously, indicates that the human extracted strategic physical rules are highly generalizable. Therefore,  we foresee several areas of improvement: 1) agents should learn and store generalizable abstract causal knowledge \cite{marcus2020}, e.g. strategic physical rules. For example, humans understand not only shooting a bird at a pig can destroy the pig, but also the pig is destroyed because when the bird hits the pig, a force is exerted by the bird onto the pig \cite{leslie_1994} and if the force is large enough, an object will be destroyed. One possible way to learn this abstract causal knowledge is through Explanation-Based Learning (EBL) \cite{EBL}, where an agent constructs an explanation for initial exemplars and then constructs a candidate rule that depends only on the explanation; if the rule is proven true for a small number of additional exemplars, the rule is adopted. As the representation of abstract and causal knowledge allows for symbolic manipulation \cite{marcus2020}, 2) it is also worthwhile to explore the possibility of combining deep learning techniques with reasoning over knowledge systems in physical domains, where Neural-Symbolic methods such as NS-DR \cite{CLEVRER} and NS-CL \cite{DBLP:journals/corr/abs-1904-12584} have shown promising results on physical reasoning.

Phy-Q can be advanced in different directions. Characteristics such as deforming can be introduced to the objects in the tasks. Further, complex scenarios can be added to the testbed by combining the existing scenarios. This will also enable the measuring of the combinatorial generalization of the agents. Moreover, additional physical scenarios that are not covered in the testbed can be introduced such as shape constancy, object permanence, spatiotemporal continuity, and causality. We hope that Phy-Q can provide a foundation for future research on developing AI agents with human-level physical reasoning capabilities, thereby coordinating research efforts towards ever new goals.

\section{Methods}

\subsection{Phy-Q Testbed Tasks and Evaluation}

In this section, we discuss the details of the designing of task templates and the generation of task instances. We also explain the evaluation settings we have used in the testbed.
 
\subsubsection{Task Templates and Task Generation}

We design task templates in Angry Birds for each of the 15 physical scenarios mentioned above. A task template can be solved by a specific strategic physical rule and all the templates belonging to the same scenario can be solved by the high-level strategic physical rules discussed above. 
To guarantee this, in the Phy-Q testbed we handcrafted the task templates as existing task generators for Angry Birds \cite{StephensonAIBIRDSLevel, deceptive_generator} do not generate tasks according to a strategic physical rule. Also, we ensure that if an agent understood the strategic physical rule to solve the template, they can solve the template without requiring highly accurate shooting, e.g., the template can be solved by shooting at a specific object rather than shooting a specific coordinate. This design criterion is followed to reduce the dexterity requirement when solving the tasks in our testbed. We have developed 2-8 task templates for each scenario totalling 75 task templates. Figure 1 shows example task templates for the 15 scenarios.

We generate 100 game levels from each template and we refer to these game levels as tasks of the task template. All tasks of the same template share the same strategic physical rule to solve. Similar to \cite{Phyre}, the tasks are generated by varying the location of the game objects in the task template within a suitable range. Furthermore, various game objects are added at random positions in the task as distractions, ensuring that they do not alter the solution of the task. When generating the tasks each task template has constraints to satisfy such that the physical rule of the template is preserved. For example, the constraints can be, which game objects should be directly reachable by a bird shot from the slingshot, which game objects should be unreachable to the bird, which locations in the game level space are feasible to place the game objects, etc. These constraints are specific to each task template; they were determined by the template developers and hard coded in the task generator.

\begin{figure}[t]
  \centering
  \includegraphics[width=0.98\linewidth]{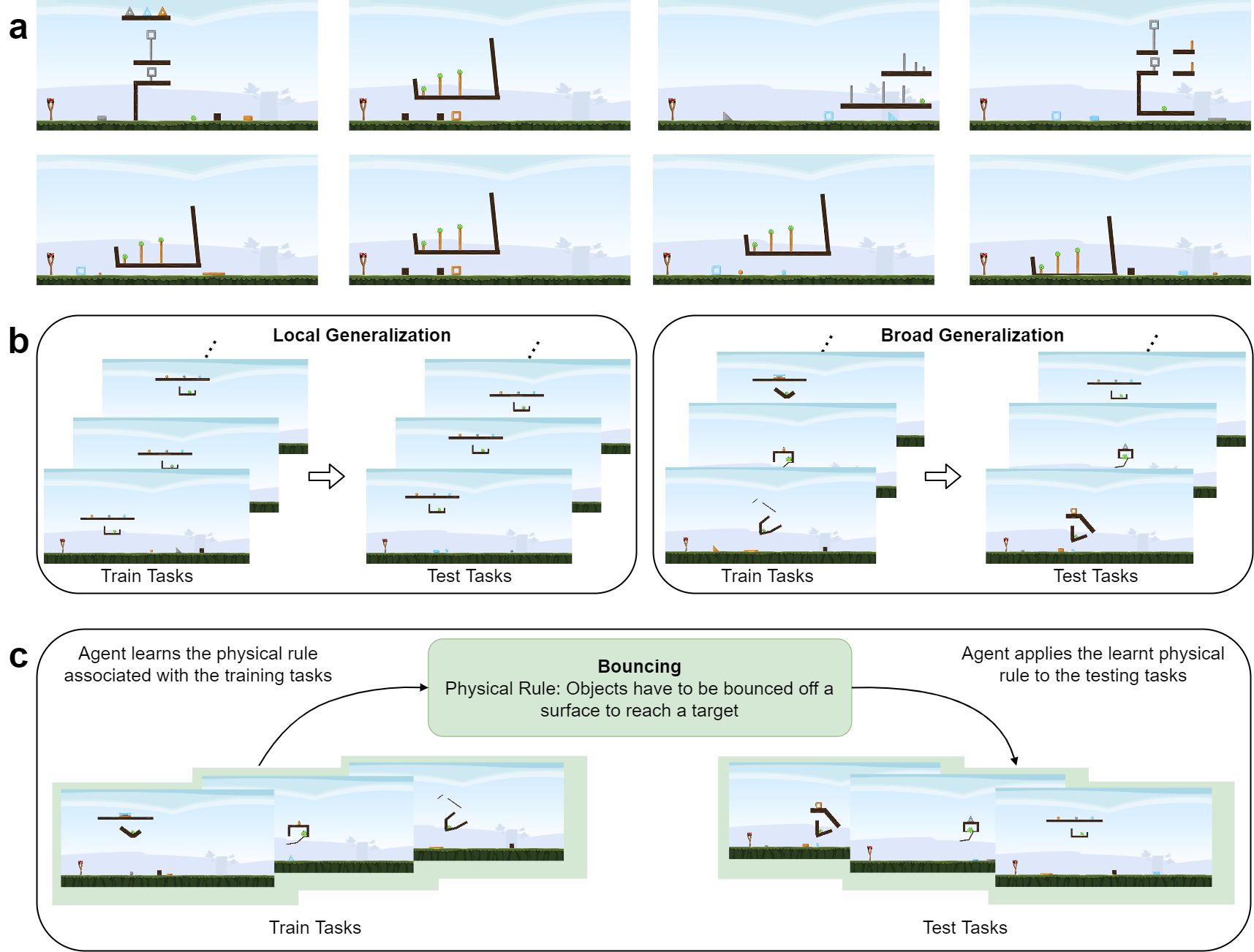}
  \caption{Phy-Q task templates and evaluation settings. a) The first row shows the task templates of the relative height scenario and the second row shows the tasks generated using the second task template in the first row. b) The \textit{local generalization} and the \textit{broad generalization} evaluation settings. c) An illustration of how generalizing a physical rule is evaluated in the \textit{broad generalization} evaluation using the bouncing scenario as an example.
  }
  \label{fig_evaluation_setup}
\end{figure}

Although we provide 100 tasks for each task template, we also provide a task variation generation module to generate more tasks if needed. Figure 2a
shows task templates of the relative height scenario and example tasks generated from a single task template. All the 75 task templates and example task variations can be found in Supplementary C.

\subsubsection{Proposed Evaluation Settings} \label{section Evaluation Settings}

The spectrum of generalization proposed by Chollet \cite{Chollet19} can be used to measure intelligence as laid out by theories of the structure of intelligence in cognitive psychology. There are three different levels in the spectrum: \textit{local generalization}, \textit{broad generalization}, and \textit{extreme generalization}. Having 15 physical scenarios, a variety of task templates for each scenario, and task variations for each task template, our testbed is capable of evaluating all of the three different generalization levels. However, in this work,  we focus on measuring the \textit{local generalization} and the \textit{broad generalization} of agents, as \textit{local generalization} is the form of generalization that has been studied from the 1950s up to this day and there is an increasing research interest in achieving \textit{broad generalization} \cite{Chollet19}.

More formally, consider each scenario $scenario_i$ in the set of all scenarios $SCENARIO$, where $|SCENARIO| = 15$, we define template $template_j \in scenario_i$, where $|scenario_i| = NT_i$ and $NT_i$ is the number of templates we included for $scenario_i$. As we have $100$ tasks for each templates, we define $task_k \in template_j$, where $|template_j| = 100$ for all templates. i.e., each scenario is a set of tasks and the tasks in a scenario are partitioned into templates. 

To evaluate local generalization within a particular template, we train an agent on some (80\% in practice) of the tasks in a template and evaluate it on the remaining tasks of the same template. To evaluate broad generalization within a particular scenario, we train an agent on the tasks of some of the templates of that scenario and evaluate it on the tasks of the other templates of the same scenario (See Supplementary E for the division of task templates for training and testing for each scenario). 

We evaluate the \textit{broad generalization} performance for all 15 scenarios. We assume that if an agent learns the required strategic physical rule to solve a set of task templates, it should be able to apply the same strategic physical rule to solve unseen tasks from other templates within the same scenario. As opposed to this, the performance on \textit{local generalization} evaluation may not represent an agent's physical rule generalizing capability but memorizing a special-purpose heuristic. Figure 2b 
is a diagrammatic representation of the two evaluation settings and Figure 2c 
shows an illustration of how generalizing a physical rule is evaluated in the \textit{broad generalization} evaluation setting.

Our physical reasoning quotient (Phy-Q) is inspired by the deviation IQ \cite{wechslerIQ} of humans.
We calculate Phy-Q of an agent using the results of our \textit{broad generalization} evaluation since we consider that this evaluation measures the agent's ability in generalizing strategic physical rules. 
When calculating the Phy-Q we exclude the first two scenarios: single force and multiple forces, as the solution for the two scenarios is directly shooting the bird to the exact location of the pig. Given that we have provided a trajectory planner for both humans and agents, solving the tasks of the two scenarios is straightforward. This is also evident from the exceptionally high results (in Section \nameref{loc_and_bro_results}) of the Pig Shooter agent that directly shoots at the pigs without doing any physical reasoning. We define the Phy-Q score as follows:

\begin{equation}
\resizebox{0.9\linewidth}{!}{
\textit{$Z_{agent}$} = $\frac{1}{|SCENARIO - \{scenario_1,scenario_2\}|} \sum_{m=3}^{|SCENARIO|} \frac{P_{agent,m} - {P_{human,m}}}{\sigma_{human,m}} $}
\end{equation}

\begin{equation}
\resizebox{.42\linewidth}{!}{
\textit{Phy-Q score$_{agent}$} = $100 + Z_{agent} \frac{100}{|Z_{random}|}$
}
\end{equation}

where $P_{n,m}$ is the average pass rate of subject \textit{n} in the $m^{th}$ scenario. $\sigma_{human,m}$ is the standard deviation of human pass rate in scenario $m$ and $random$ represents the random agent which selects a random action (See Section \nameref{baseline agent}). A Phy-Q score of 100 represents the agent having an average human level performance and if the score is less than 100, the agent's performance is less than the average humans' performance and vice versa. As the random agent does not have any physical reasoning capabilities, we bring the random agent's Phy-Q score to zero. 
Therefore, we set the scaling factor to 100/$|Z_{random}|$ which is 13.58 as compared to 15 of IQ.
Therefore, a Phy-Q score of more than zero indicates performance better than an agent that selects a random action.

\subsection{Experiments}

We conduct experiments on baseline learning agents to measure how well they can generalise in two different settings: \textit{local generalization} and \textit{broad generalization}.
We also conduct experiments using heuristic baseline agents in the two generalization settings. In addition, we establish human performance in the 15 scenarios. 
Further, we conduct an additional experiment using heuristic agents in AIBIRDS competition game levels to examine if the performance of agents in the testbed resembles the performance in the competition. 

\subsubsection{Baseline Agents} \label{baseline agent}

We present experimental results of nine baseline agents: two DQN agents - one uses screenshot as input and the other uses symbolic representation, two relational agents - one uses screenshot as input and the other uses symbolic representation, four heuristic agents from the AIBIRDS competition, and a random agent.

\paragraph{Learning Agents}
For learning agents, we tested value-based and policy-based (see Supplementary I) reinforcement learning algorithms and report the results of Double Dueling DQN agents and Relational DQN agents.

\begin{itemize}[leftmargin=*]
    \item \textbf{Deep Q-network (DQN): } The DQN \cite{mnih2015humanlevel} agent collects state-action-reward-next state quadruplets at the training time following decaying epsilon greedy.   
    We define the reward function as task pass status, meaning the agent receives $1$ if the task is passed and $0$ otherwise. 
    We report the performance of Double Dueling DQN \cite{DBLP:journals/corr/WangFL15, DBLP:journals/corr/HasseltGS15} two different input types: symbolic representation and screenshot. We refer to the Dueling Double DQNs we experimented with the two different input types as D-DDQN-Image and D-DDQN-Symbolic.
    
    \item \textbf{Relational Deep Q-network}: The relational agent consists of the relational module \cite{Zambaldi2018RelationalDR} that was built on top of the deep Q-network. The aim of this agent is to generalize over the presented templates/events by using structured perception and relational reasoning. In our experiments, we wanted to test if the relational agent would be able to learn the important relations between the objects that could be generalized to other templates or events. We have tested the agent with symbolic and image input types and refer to them as Relational-symbolic and Relational-image agents respectively.

\end{itemize}

\paragraph{Heuristic Agents:} 

The heuristic agents are based on hard-coded strategic physical rules designed by the developers. We included four heuristic agents from AIBIRDS competition.
We compare heuristic agents' performance on our testbed with the generalization performance of the baseline learning agents.

\begin{itemize}[leftmargin=*]
    \item \textbf{Bambirds:} Bambirds is the winner of the 2016 and 2019 AIBIRDS competitions. The agent chooses one of nine different strategies. The strategies include creating a domino effect, targeting blocks that support heavy objects, maximum structure penetration, prioritizing protective blocks, targeting pigs, and utilizing certain bird’s powers \cite{Bambirds}.
    \item \textbf{Eagle's Wing:} Eagle’s Wing is the winner of the 2017 and 2018 AIBIRDS competitions. This agent selects action based on strategies including shoot at pigs, destroy most blocks, shoot high round objects, and destroy structures \cite{EagleWings}.
    \item \textbf{Datalab:} Datalab is the winner of the 2014 and 2015 AIBIRDS competitions. The agent use strategies: destroy pigs, destroy physical structures, and shoot at round blocks. The agent selects the strategy based on the game states, possible trajectories, bird types, and the remaining birds \cite{Datalab}.
    \item \textbf{Pig Shooter:} The strategy of the Pig Shooter is to directly shoot at the pigs. The agent shoots the bird on the slingshot by randomly selecting a pig and a trajectory to shoot the pig \cite{Stephenson2018TheCompetition}.
\end{itemize}

\paragraph{Random agent:} For each shot, the agent selects a random release point $(x, y)$, where $x$ is sampled from $[-100,-10]$ and $y$ from $[-100,100]$ relative to the slingshot. It also provides a tapping time that is when the bird is in between $50\%$ and $80\%$ of the trajectory length where applicable.

\subsubsection{Experimental Setups} \label{section experimental setup}
\paragraph{Human experiment setup:} Experiments were approved by the Australian National University committee on human ethics under protocol 2021/293. Participation was voluntary and they were not compensated monetarily.
The volunteers represented males and females and were around the age range 18-35. They were not experienced Angry Birds players. Participants provided consent to use their playdata. 
For each of them, we provided two tasks from each physical scenario for the 15 scenarios in Phy-Q (except the manoeuvring scenario which used four tasks representing the four types of birds with powers). We provide a trajectory visualizer of the bird to the participants to remove the need for precise shooting. If the participants solved a task or failed to solve a task in five attempts, they move to the next task.
As humans acquire physical reasoning capabilities from their infancy \cite{rolling_falling,falling}, using an evaluation setting we proposed for agents does not exactly measure the generalization ability of humans. Therefore, we measure the task performance in humans using the pass rate. 

\paragraph{D-DDQN and Relational DQN experimental setup:} \label{DQN_exp_setup}
We conduct separate experiments on the D-DDQN and Relational agents in the two settings: \textit{local generalization} and \textit{broad generalization} setting.
For the \textit{local generalization} evaluation, we run 10 sampling agents that use the same DQN model to collect experiences. Each sampling agent runs on the randomly selected task for 10 episodes. After the set of experiences is collected, the DQN model is trained for 10 epochs with a batch size of 32. We train DQN until it either converges or reaches $N$ update steps, where $N$ is the number of training tasks per template divided by 5. Similar to \cite{Phyre}, for each batch, we sample 16 experiences that a task is solved and 16 that failed. We train our agent on 80$\%$ of the tasks of the task template and evaluate on the rest of the tasks of the same template. We used the same training setting for all of the task templates. At the testing time, the agent runs on each of the testing tasks only once and selects the action that has the highest Q-value for a given state.
For {\it broad generalization} evaluation, we use the same training and testing setting as in the \textit{local generalization} evaluation, except we train our agents on the tasks in the training templates in each scenario and test on the tasks from the testing templates.

\paragraph{Heuristic agents experimental setup:} We conduct two experiments using the AIBIRDS heuristic agents. The first experiment is to evaluate the \textit{local and broad generalization} capabilities and the second is to evaluate the performance in the AIBIRDS competition game levels.

\begin{itemize}[leftmargin=*]
    \item \textbf{\textit{Local and broad generalization setup:}} 
    Due to the randomness in the heuristic agents, we allow them to have five attempts per task and calculate the task pass rate by averaging the result over these five attempts. For the \textit{local generalization} setting, the agents were tested on the same $20\%$ of the test tasks from each task template (1500 tasks in total) used for the D-DDQN evaluation. We report the \textit{local generalization} performance by averaging the pass rates of all templates. For the \textit{broad generalization} setting, the same testing templates used for the D-DDQN evaluation were used and the within scenario pass rate is calculated by averaging over all the tested templates within the scenario.

    \item \textbf{AIBIRDS competition setup:} We evaluate the AIBIRDS heuristic agents on 2021 AIBIRDS competition game levels to compare their performance in the competition game levels and the Phy-Q testbed tasks. We exclude the competition game levels with unrealistic effects as our focus in the testbed is scenarios with realistic physics. The game levels used for this evaluation are shown in Supplementary G. In the AIBIRDS competition, the agent with the highest score wins the competition. Therefore in this experiment, we record the score and pass rate of the agents. The agents are allowed to have five attempts per game level to account for their randomness. Altogether an agent had 40 plays.
    
\end{itemize}

\paragraph{Random agent experimental setup:}
The random agent was tested on the same testing tasks set from each task template. We run the random agent 50 times per task and report the average pass rate of these 50 attempts.
Same as how we evaluate the heuristic agents, we further average the task performance within the same task template and average the pass rate of all the templates to present the \textit{local generalization} performance. For the \textit{broad generalization} setting, within scenario pass rate is calculated by averaging over all the tested templates within the scenario.

\subsection{Results and Analysis} \label{result and analysis}

In this section, we first present and analyze the results obtained from our experiment with human players. Next, we present the results obtained from our experiments in measuring the local and broad generalization ability of agents and Phy-Q score. We further analyze the results and discuss what we can derive from the experiments. We also discuss the results obtained from the heuristic agents in the 2021 AIBIRDS competition levels and Phy-Q testbed tasks to show how the testbed can be used as a guide for the competition. 

\subsubsection{Human Performance}
\begin{figure}[t]
  \centering
  \includegraphics[width=0.97\linewidth]{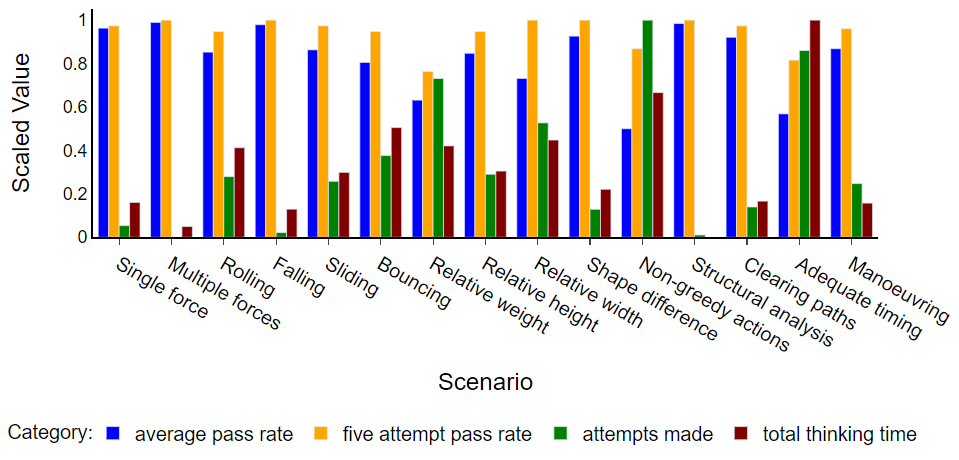}
  \caption{Performance of human players. The x-axis is the scenario and y-axis is the scaled value.}
  \label{fig_human_scaled}
\end{figure}
Figure 3 presents the average pass rate, the pass rate the human players achieved within five attempts, the maximum number of attempts made, and the total thinking time of human players for the 15 capabilities.
The average pass rate is calculated as 100\% if the player passes at the first attempt and if the player passes at the fifth attempt, the pass rate is 20\%. 
We record the thinking time of an attempt as the time between the task loading and the player making the first action. The total thinking time of a player is the sum of the thinking time of all his/her attempts. 
The number of attempts made and the total thinking time is scaled to 0-1 using min-max scaling in Figure 3.
Charts with the real values are available in Supplementary F. 

Overall, human players have passed almost all the tasks in each scenario within the five attempts. On average, they used 1.86 attempts per task and took 23.73 seconds to think per task. On average, the low number of attempts to pass the tasks shows that the dexterity required to solve the tasks when the strategy is determined is low. The average thinking time per task shows that humans have to carefully think about the strategic physical rule required to solve the task.

Humans have the longest thinking time for the tasks in the adequate timing scenario but the average pass rate for these tasks is the second lowest. Similarly, the tasks from the non-greedy actions scenario have the lowest average pass rate with the highest number of attempts while the thinking time is the second longest. This shows that figuring out the correct strategies for the tasks of these scenarios has been difficult for humans. 
In the relative weight scenario, the pass rate achieved within five attempts is the lowest. But the thinking time is average for this scenario. This suggests that some humans take the action without carefully thinking about the strategy, and the strategy realized at a glance is not the correct strategy to solve the task. This also agrees with our observation that humans are overconfident in their wrong actions. 

\subsubsection{Local and Broad Generalization Performance and Phy-Q Score}\label{loc_and_bro_results}

\begin{figure}[t]
  \centering
  \includegraphics[width=1.0\linewidth]{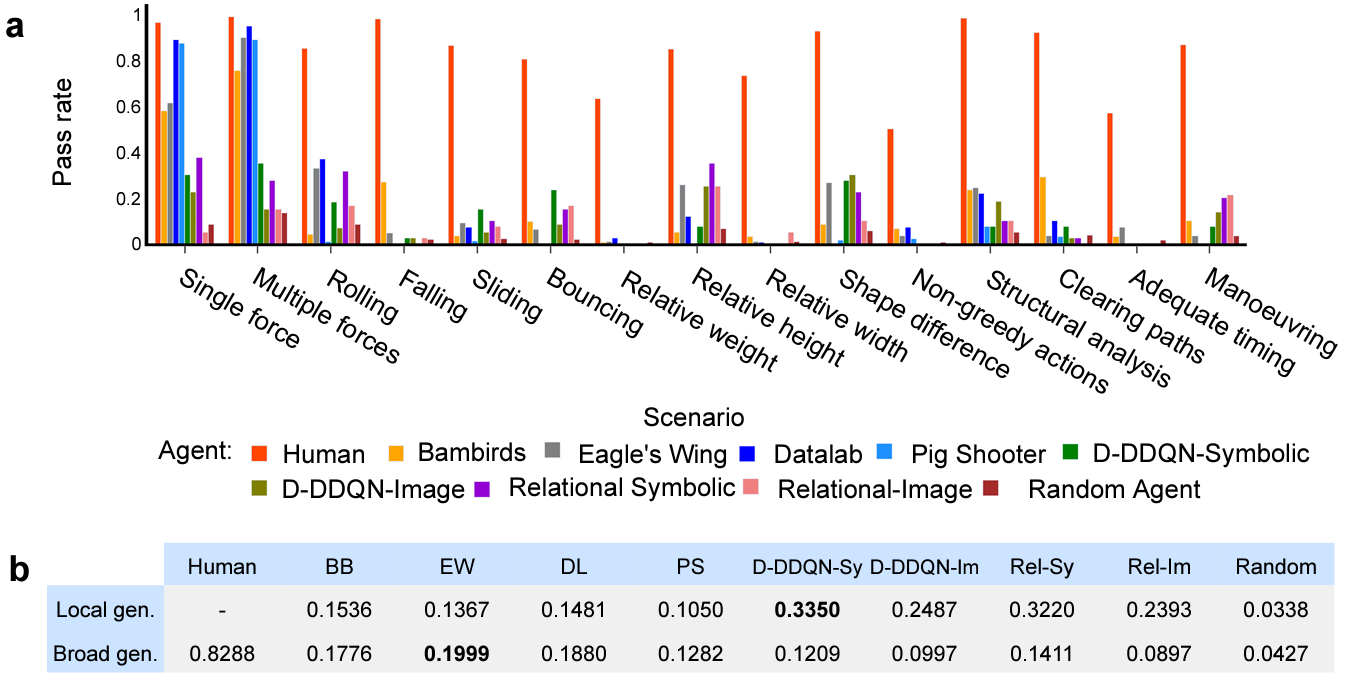}
  \caption{Generalization performance of the baseline agents. a) \textit{Broad generalization} of the baseline agents. The x-axis is the scenario and y-axis is the passing rate. b) Comparison of agents for \textit{local generalization} and \textit{broad generalization} (Phy-Q score). The results are shown for the nine agents: Bambirds (BB), Eagle's Wing (EW), Datalab (DL), Pig Shooter (PS), D-DDQN-Symbolic (D-DDQN-Sy), D-DDQN-Image (D-DDQN-Im), Relational-Symbolic (Rel-Sy), Relational-Image (Rel-Im), and Random. The performance of the best performing agent is shown in bold. Learning agents have higher \textit{local generalization} values and lower values in \textit{broad generalization} than the performance of heuristic agents. Human performance is way beyond agents.}
  \label{fig_broad_generalization}
\end{figure}

\paragraph{\textit{Local generalization} performance}

Figure 4b first row presents the average \textit{local generalization} evaluation pass rate for all of our baseline agents, we have also included the full results of pass rate per agent per template in Supplementary D. The table shows that the four learning agents perform significantly better than their heuristic counterparts. While both the symbolic learning agents and both the image learning agents on average pass approximately $33\%$ and $24\%$ of the test levels respectively, the previous champions in AIBIRDS competition-Bambirds, Eagle's Wing, and Datalab-pass around only half of the levels as compared with the learning agents, averaging $15\%$, $14\%$, and $15\%$ respectively. This agrees with what is generally accepted that deep learning systems can perform a single narrow task much better than heuristic methods when enough densely sampled training data is available. 

\paragraph{\textit{Broad generalization} performance}

Figure 4a presents the average pass rate of test templates of the \textit{broad generalization} evaluation of all the baseline agents and human players. It is clear that the humans substantially outperform all the other agents, while all the agents have above-chance performance compared to the random agent. Heuristic agents achieved a better pass rate in the single force scenario (scenario 1) and multiple force scenario (scenario 2) as these two scenarios are corresponding to the essential ability needed to play Angry Birds - shooting directly at pigs. 
It can be seen that the heuristic agents generally perform better if the physical scenario is covered in their built-in rules. For example, Datalab and Eagle's Wing have a built-in strategic physical rule to roll round objects and they have the highest pass rate in scenario 3 (rolling) among all agents. For scenario 4 (falling) and scenario 13 (clearing paths), Bambirds dominates the leaderboard of pass rate because it explicitly analyses spatial relationships between blocks and pigs and is the only heuristic agent with the `prioritizing protective blocks' rule. 

The second row in Figure 4b shows the overall average pass rate for the \textit{broad generalization} evaluation of the agents and humans. The heuristic agents' results were obtained in a similar way as the \textit{local generalization} evaluation, except we only consider the tasks from the testing task templates given to the learning agents. In contrast to the \textit{local generalization} results, in this evaluation setup, the learning agents have worse results than all the heuristic agents. The D-DDQN-Symbolic and the D-DDQN-Image agents have an average pass rate of 12\% and 10\% respectively while Relational-Symbolic and Relational-Image have 14\% and 9\% respectively. The champions in the AIBIRDS competition have an almost double pass rate compared to the learning agents. Our result further advocates the claim that deep learning agents often exploit spurious statistical patterns instead of learning in a meaningful and generalizable way as humans do \cite{Chollet19,EBL,bengio2020,Yixin2020,marcus2020}.

\paragraph{Phy-Q score}

\begin{figure}[t]
  \centering
  \includegraphics[width=0.8\linewidth]{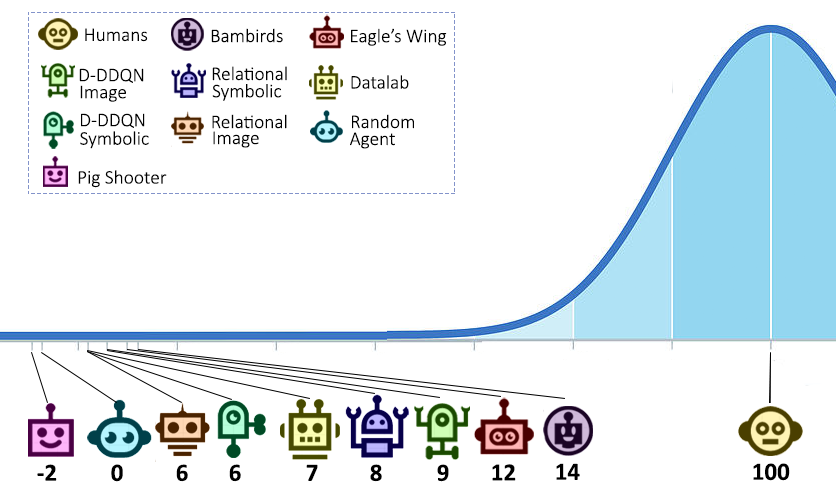}
  \caption{Phy-Q Score of human players and the agents. Phy-Q score of human players is at 100 and all agents are far below human players.}
  \label{fig_phy_q_distribution}
\end{figure}

As discussed in Section \nameref{section Evaluation Settings}, the Phy-Q score of humans is set to 100 while the random agent is set to 0. A Phy-Q score above 100 indicates superhuman performance. 
Figure 5 shows the positions of agents and humans in the Phy-Q score distribution. Even though Eagle's Wing was the first in the \textit{broad generalization} leader board (even after removing the results of the first two scenarios Eagle's Wing: 0.1142, Bambirds: 0.1022), Bambirds has taken the lead in the Phy-Q score dragging Eagle's Wing into the second place. This is because the Phy-Q score positions the agent with respect to human performance.
Interestingly D-DDQN-Image agent and Relational-Symbolic agents have higher Phy-Q values compared to Datalab. 
Similar to the above reason, this is due to the positioning of the agents with respect to human performance. 
Moreover, the Phy-Q of the Pig Shooter is negative. This result is expected as the Pig Shooter only shoots at the pigs, thus having a below chance performance compared to the random agent. Overall, it can be seen that all the agents are far below the humans' Phy-Q score. 

\subsubsection{AIBIRDS Competition Performance}

\begin{table}
    \scriptsize
    \centering
    \caption{Comparison of AIBIRDS competition performance with Phy-Q score of heuristic agents}
    \label{table AIBIRDS competition results}
    \begin{tabular}{c|cccc}
    \toprule
        Rank from Phy-Q score & 1 & 2 & 3 & 4 \\
         \midrule
        Agent Name & Bambirds & Eagle's Wing & Datalab & Pig Shooter\\
        \midrule
        Competition performance & 0.2727 & 0.2000 & 0.2000 & 0.0000 \\
        \midrule
        Phy-Q score & 14 & 12 & 7 & -2\\
        \bottomrule
    \end{tabular}
\end{table}

Table \ref{table AIBIRDS competition results} presents the results of the AIBIRDS heuristic agents in the AIBIRDS competition game levels. 
As can be seen from the results, the pass rate of the agents in competition game levels agree to the rank they achieved using the Phy-Q score. Eagle's Wing and Datalab have achieved the same pass rate of 0.2. However, considering the total score, Eagle's Wing has obtained 667,394 while Datalab has obtained 634,435 taking Eagle's Wing to the second position. 
Overall, these results portray that the tasks in the Phy-Q testbed are representative of the tasks in the AIBIRDS competition.

From this result, we infer that the physical scenarios available in the Phy-Q tasks are also the scenarios that are commonly encountered in the AIBIRDS competition game levels. The within scenario (\textit{broad generalization}) evaluation we have conducted can be used to identify an agent's ability in performing in those individual physical scenarios. Therefore one can use the Phy-Q testbed to thoroughly analyze the physical reasoning capabilities of their AIBIRDS agent and figure out where the agent falls short and improve on those capabilities. Additionally, the human performance results we have established in the 15 scenarios facilitate the comparison of the agents' performance with humans' performance, allowing us to set targets for agents to achieve human level performance in those scenarios. Thus, the Phy-Q testbed and the evaluation settings we proposed in the testbed can be used to better evaluate AIBIRDS agents and guide them towards achieving human level performance in the competition.

\section*{Data Availability}
The data collected from human players and baseline agents have been made available at \url{https://github.com/phy-q/benchmark/tree/master/playdata}. 

\section*{Code Availability}
The testbed software and baseline agents' codes have been made available at \url{https://github.com/phy-q/benchmark} \cite{phyqGitRepo}.

\section*{Acknowledgements}
This work is supported by the Defense Advanced Research Projects Agency (DARPA) and the Army Research Office (ARO) and was accomplished under Cooperative Agreement Number W911NF-20-2-0002 (received by J.R and P.Z). 
We convey our gratitude to all volunteers that participated to play the game and the anonymous reviewers for their constructive input. We thank Alexander Yang for fruitful discussions.  

\section*{Contributions}
C.X., V.P., C.G., and J.R. conceived this study. C.X., V.P., and C.G. wrote the manuscript. C.X. implemented learning agents, V.P. designed task templates and analysed the data, C.G. designed task templates and developed the task generator. C.X. conducted the experiments on learning agents and C.X., E.N. wrote the section on learning agents. C.G. and V.P. conducted experiments on non-learning agents and conducted the AIBIRDS competition evaluation. C.X. and P.Z. adapted the Angry Birds framework to the testbed. J.R. provided feedback and supervision.

\section*{Competing interests}
The authors declare no competing interests.

\printbibliography

\newpage

\appendix
\section*{Appendix}

\section{The Objects in Phy-Q Tasks}

\begin{figure}[h!]
  \centering
  \includegraphics[width=0.34\linewidth]{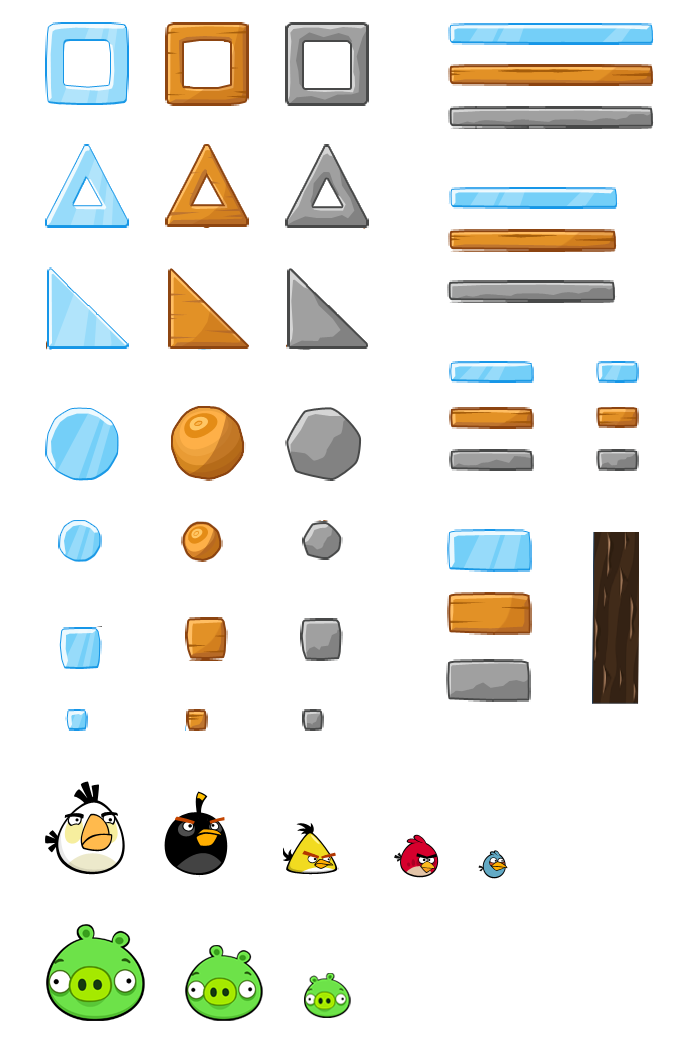}
  \caption{Objects that appear in the Phy-Q tasks reproduced from Angry Birds [1]. The shape and the size of the objects are fixed, except the platform object (dark brown) which can appear in different shapes and sizes.}
  \label{fig_objects_in_AB}
\end{figure}

\begin{table}[h!]
\small
  \caption{Powers of the birds in the tasks. The birds with powers are used only for the tasks in the manoeuvring scenario. The powers can be activated by tapping the bird in flight.}
  \label{bird_special_powers}
  \centering
  \begin{tabular}{ll}
    \toprule
    Bird & Power\\
    \midrule
    Red bird & No power\\
    Yellow bird & Accelerates forward\\
    Blue bird & Splits into three birds\\
    White bird & Drops an explosive egg\\
    Black bird & Explodes itself\\
    \bottomrule
  \end{tabular}
\end{table}

\newpage
\section{Phy-Q Task Representation}
\begin{figure}[h!]
  \captionsetup[subfigure]{labelformat=empty}
  \centering
  \begin{subfigure}[b]{0.4\columnwidth}
    \includegraphics[width=\linewidth]{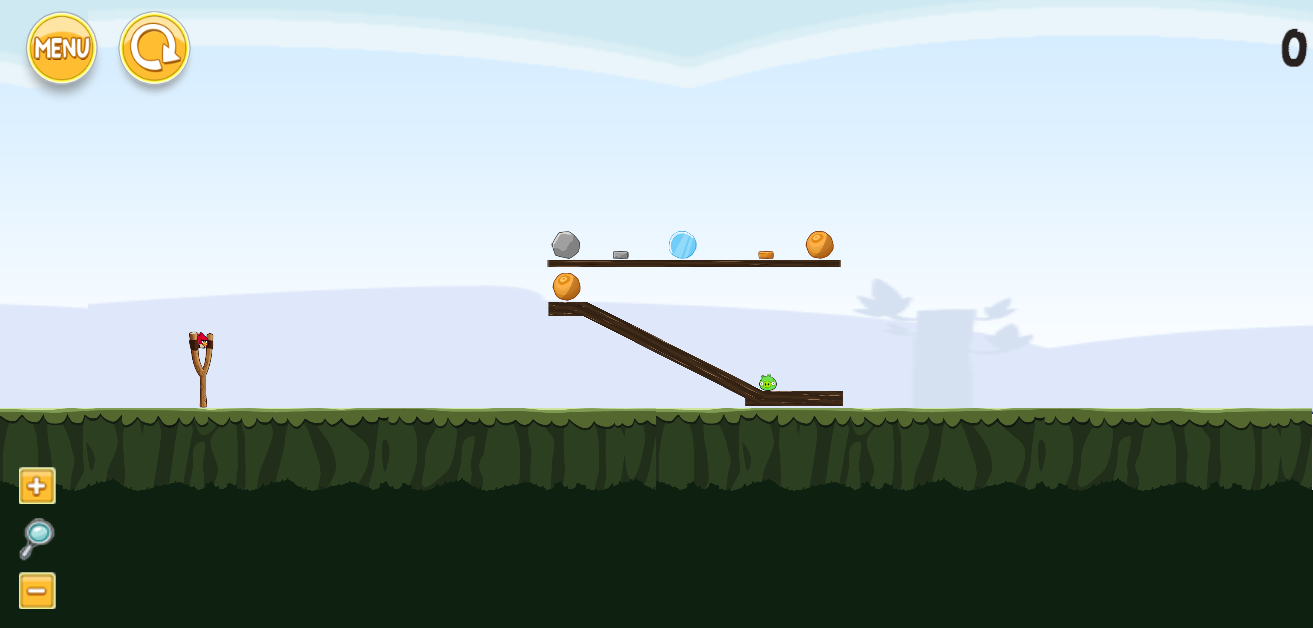}
    \caption{An example screenshot}
  \end{subfigure}
  
  \begin{subfigure}[b]{0.55\columnwidth}
    \includegraphics[width=\linewidth]{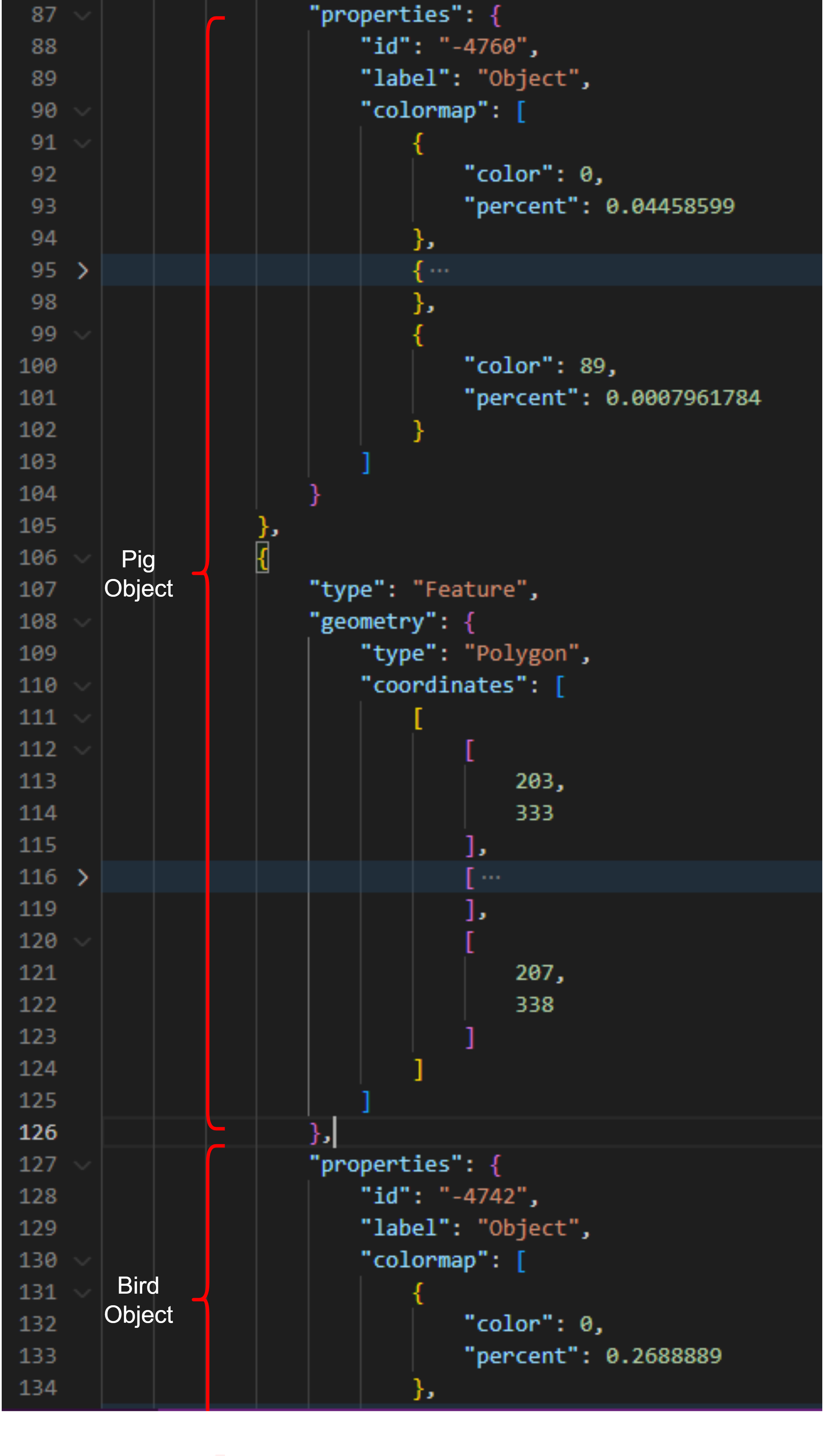}
    \caption{The corresponding symbolic representation}
  \end{subfigure}
\caption{A screenshot of an example task in Phy-Q with the corresponding symbolic representation in JSON format.}
\label{symbolic_representation}
\end{figure}

 An agent can request screenshots and/or a symbolic representation of the game level (task) at any time while playing. A game screenshot is a 480 x 640 coloured image and the symbolic representation is in JSON format containing all objects in the screenshot represented as a polygon of its vertices (provided in order) and its respective colour map. The colour map provides the list of 8-bit quantized colours that appear in the game object with their respective percentages. Figure \ref{symbolic_representation} shows an example game screenshot and the corresponding symbolic representation (part of it). 
 
\newpage
\section{Phy-Q Tasks}
\subsection{Task Templates in Phy-Q}
\begin{figure}[H]
  \captionsetup[subfigure]{labelformat=empty}
  \centering
  \begin{subfigure}[b]{0.32\columnwidth}
    \includegraphics[width=\linewidth]{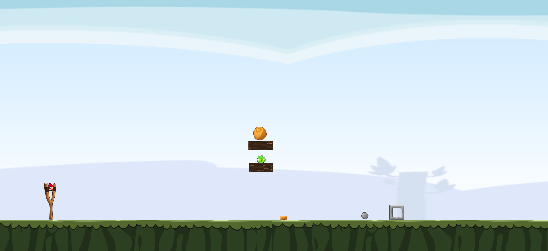}
    \caption{1.1}
  \end{subfigure}
  \begin{subfigure}[b]{0.32\columnwidth}
    \includegraphics[width=\linewidth]{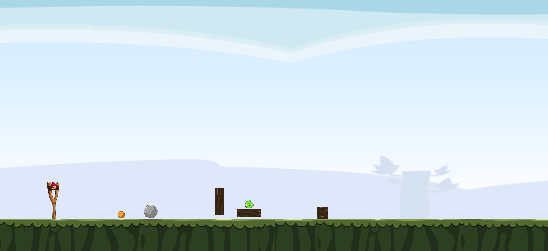}
    \caption{1.2}
  \end{subfigure}
  \begin{subfigure}[b]{0.32\columnwidth}
    \includegraphics[width=\linewidth]{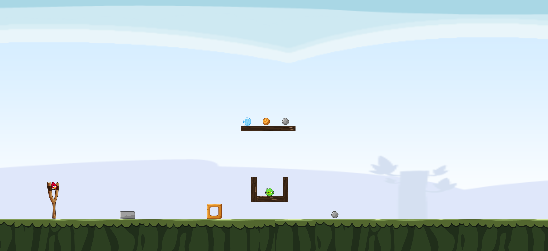}
    \caption{1.3}
  \end{subfigure}
  \begin{subfigure}[b]{0.32\columnwidth}
    \includegraphics[width=\linewidth]{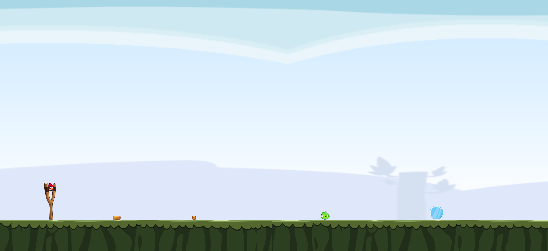}
    \caption{1.4}
  \end{subfigure}
  \begin{subfigure}[b]{0.32\columnwidth}
    \includegraphics[width=\linewidth]{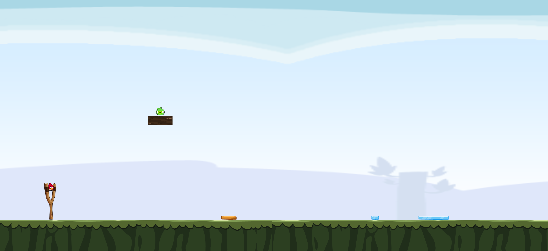}
    \caption{1.5}
  \end{subfigure}
\caption{Task templates from the single force scenario. The pig in these templates can be destroyed by directly shooting the bird at the pig.}
\label{event_1}
\end{figure}

\begin{figure}[H]
  \captionsetup[subfigure]{labelformat=empty}
  \centering
  \begin{subfigure}[b]{0.32\columnwidth}
    \includegraphics[width=\linewidth]{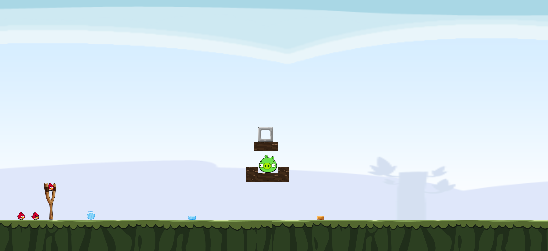}
    \caption{2.1}
  \end{subfigure}
  \begin{subfigure}[b]{0.32\columnwidth}
    \includegraphics[width=\linewidth]{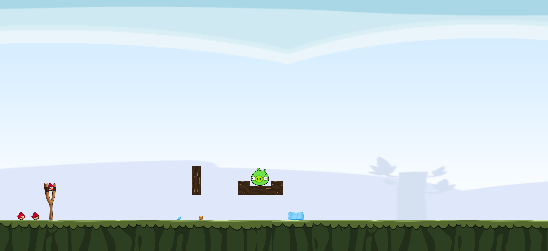}
    \caption{2.2}
  \end{subfigure}
  \begin{subfigure}[b]{0.32\columnwidth}
    \includegraphics[width=\linewidth]{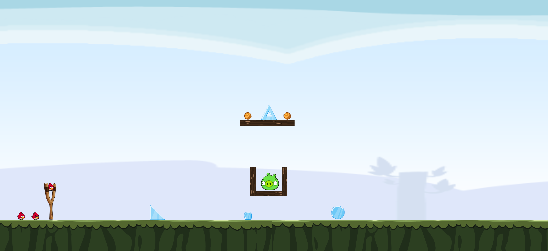}
    \caption{2.3}
  \end{subfigure}
  \begin{subfigure}[b]{0.32\columnwidth}
    \includegraphics[width=\linewidth]{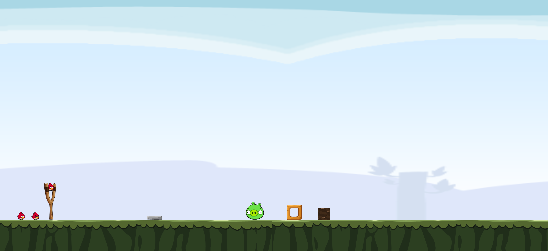}
    \caption{2.4}
  \end{subfigure}
  \begin{subfigure}[b]{0.32\columnwidth}
    \includegraphics[width=\linewidth]{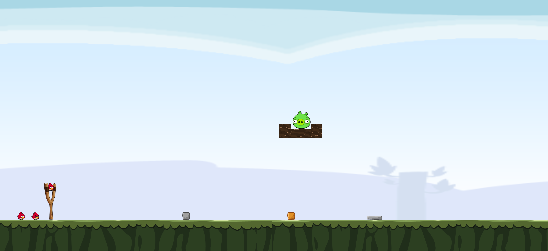}
    \caption{2.5}
  \end{subfigure}
  
\caption{Task templates from the multiple forces scenario. The pig in these templates can be destroyed by shooting multiple birds at the pig.}
\label{event_2}
\end{figure}

\begin{figure}[H]
  \captionsetup[subfigure]{labelformat=empty}
  \centering
  \begin{subfigure}[b]{0.32\columnwidth}
    \includegraphics[width=\linewidth]{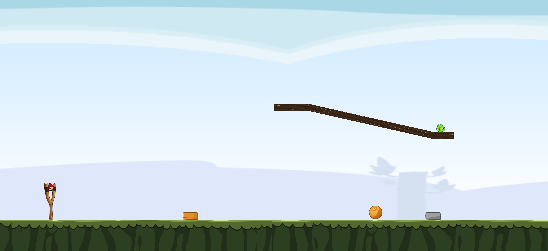}
    \caption{3.1}
  \end{subfigure}
  \begin{subfigure}[b]{0.32\columnwidth}
    \includegraphics[width=\linewidth]{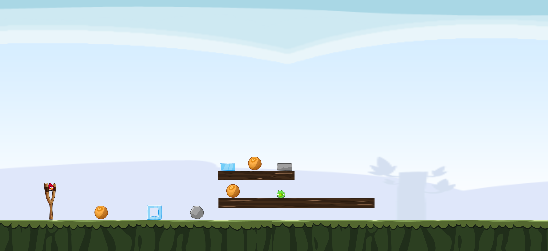}
    \caption{3.2}
  \end{subfigure}
    \begin{subfigure}[b]{0.32\columnwidth}
    \includegraphics[width=\linewidth]{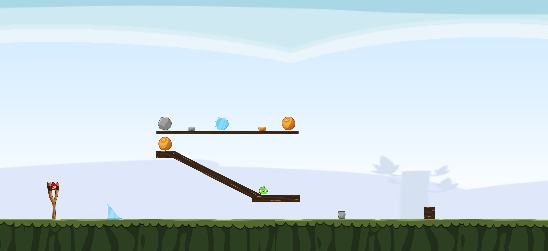}
    \caption{3.3}
  \end{subfigure}
  \begin{subfigure}[b]{0.32\columnwidth}
    \includegraphics[width=\linewidth]{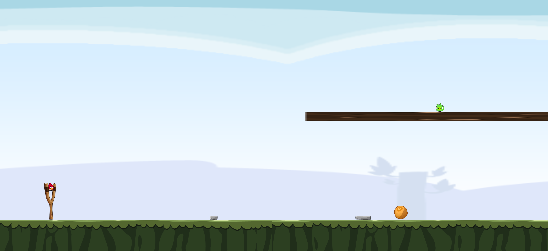}
    \caption{3.4}
  \end{subfigure}
  \begin{subfigure}[b]{0.32\columnwidth}
    \includegraphics[width=\linewidth]{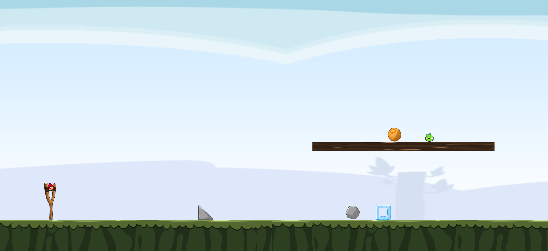}
    \caption{3.5}
  \end{subfigure}
  \begin{subfigure}[b]{0.32\columnwidth}
    \includegraphics[width=\linewidth]{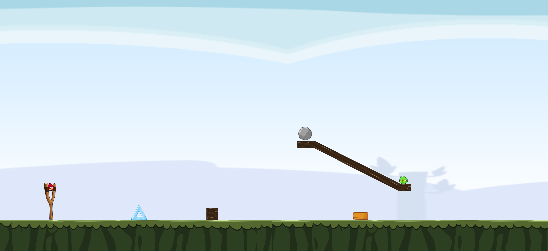}
    \caption{3.6}
  \end{subfigure}
\caption{Task templates from the rolling scenario. The pig in these templates can be destroyed by rolling the correct circular block into the pig or rolling the bird itself.}
\label{event_3}
\end{figure}

\begin{figure}[H]
  \captionsetup[subfigure]{labelformat=empty}
  \centering
  \begin{subfigure}[b]{0.32\columnwidth}
    \includegraphics[width=\linewidth]{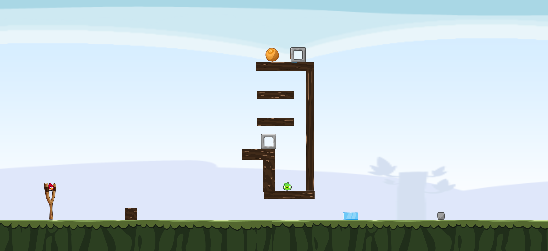}
    \caption{4.1}
  \end{subfigure}
  \begin{subfigure}[b]{0.32\columnwidth}
    \includegraphics[width=\linewidth]{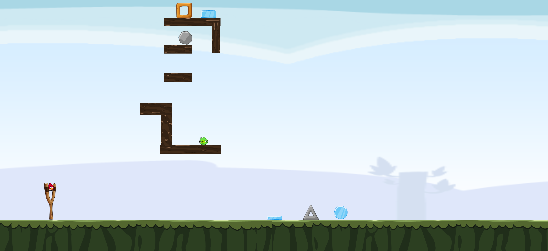}
    \caption{4.2}
  \end{subfigure}
  \begin{subfigure}[b]{0.32\columnwidth}
    \includegraphics[width=\linewidth]{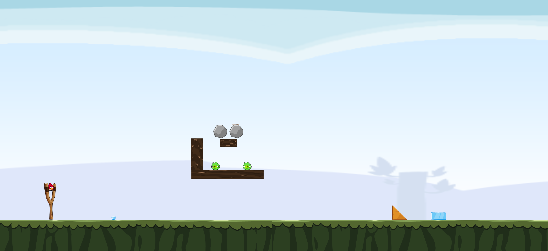}
    \caption{4.3}
  \end{subfigure}
  \begin{subfigure}[b]{0.32\columnwidth}
    \includegraphics[width=\linewidth]{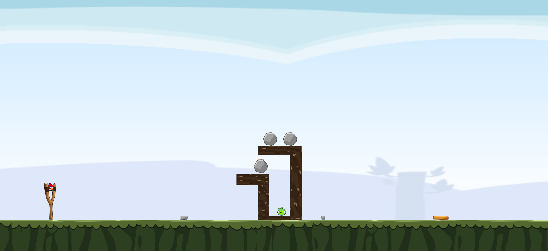}
    \caption{4.4}
  \end{subfigure}
  \begin{subfigure}[b]{0.32\columnwidth}
    \includegraphics[width=\linewidth]{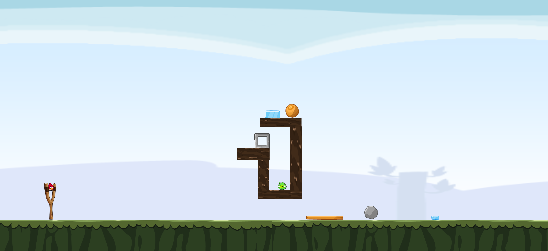}
    \caption{4.5}
  \end{subfigure}
\caption{Task templates from the falling scenario. The pig in these templates can be destroyed by falling the correct block onto the pig.}
\label{event_4}
\end{figure}

\begin{figure}[H]
  \captionsetup[subfigure]{labelformat=empty}
  \centering
  \begin{subfigure}[b]{0.32\columnwidth}
    \includegraphics[width=\linewidth]{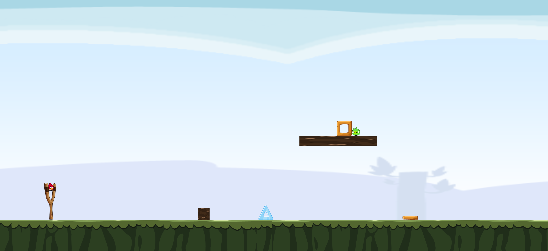}
    \caption{5.1}
  \end{subfigure}
  \begin{subfigure}[b]{0.32\columnwidth}
    \includegraphics[width=\linewidth]{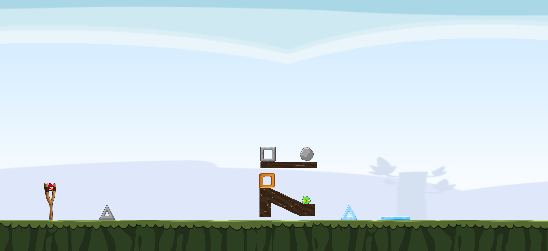}
    \caption{5.2}
  \end{subfigure}
  \begin{subfigure}[b]{0.32\columnwidth}
    \includegraphics[width=\linewidth]{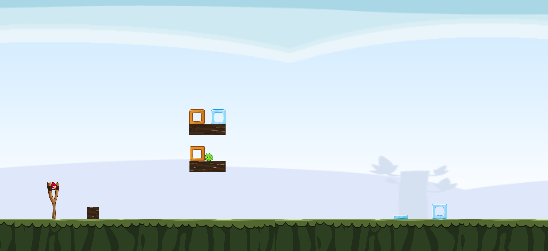}
    \caption{5.3}
  \end{subfigure}
  \begin{subfigure}[b]{0.32\columnwidth}
    \includegraphics[width=\linewidth]{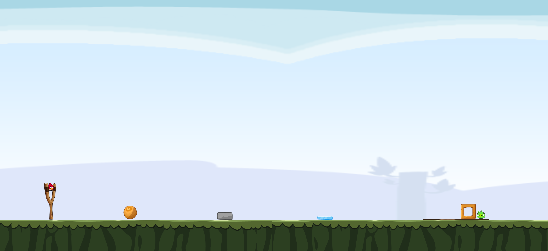}
    \caption{5.4}
  \end{subfigure}
  \begin{subfigure}[b]{0.32\columnwidth}
    \includegraphics[width=\linewidth]{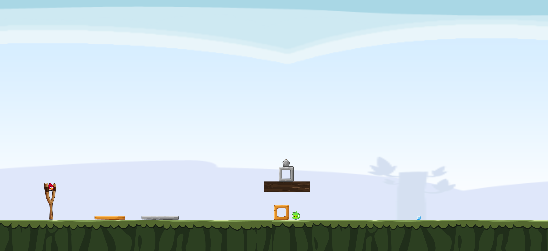}
    \caption{5.5}
  \end{subfigure}
\caption{Task templates from the sliding scenario. The pig in these templates can be destroyed by sliding the correct block into the pig.}
\label{event_5}
\end{figure}

\begin{figure}[H]
  \captionsetup[subfigure]{labelformat=empty}
  \centering
  \begin{subfigure}[b]{0.32\columnwidth}
    \includegraphics[width=\linewidth]{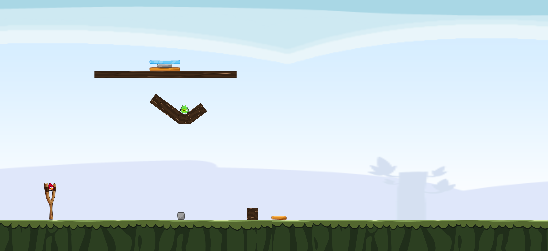}
    \caption{6.1}
  \end{subfigure}
  \begin{subfigure}[b]{0.32\columnwidth}
    \includegraphics[width=\linewidth]{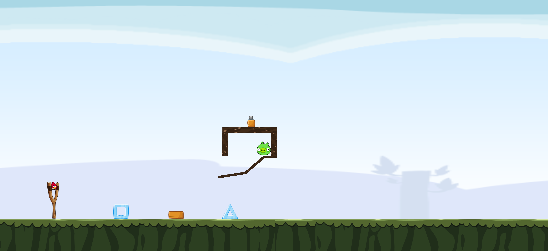}
    \caption{6.2}
  \end{subfigure}
  \begin{subfigure}[b]{0.32\columnwidth}
    \includegraphics[width=\linewidth]{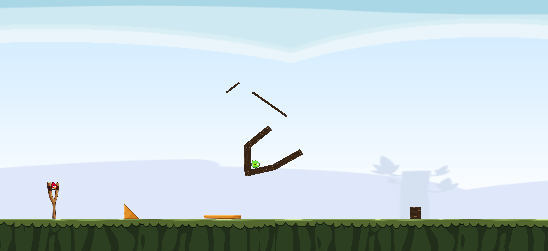}
    \caption{6.3}
  \end{subfigure}
  \begin{subfigure}[b]{0.32\columnwidth}
    \includegraphics[width=\linewidth]{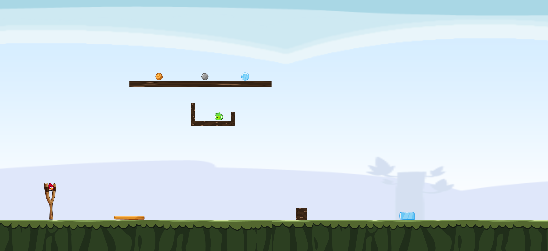}
    \caption{6.4}
  \end{subfigure}
  \begin{subfigure}[b]{0.32\columnwidth}
    \includegraphics[width=\linewidth]{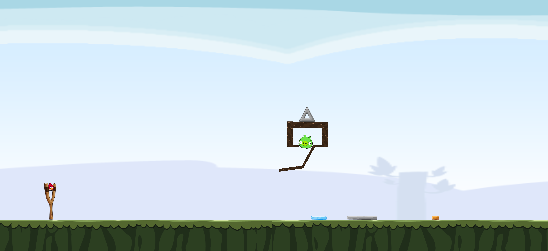}
    \caption{6.5}
  \end{subfigure}
  \begin{subfigure}[b]{0.32\columnwidth}
    \includegraphics[width=\linewidth]{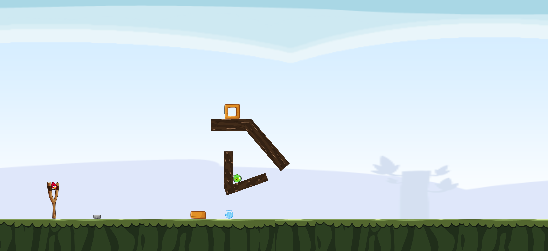}
    \caption{6.6}
  \end{subfigure}
\caption{Task templates from the bouncing scenario. The pig in these templates can be destroyed by bouncing the bird off the platforms onto the pig.}
\label{event_6}
\end{figure}

\begin{figure}[H]
  \captionsetup[subfigure]{labelformat=empty}
  \centering
    \begin{subfigure}[b]{0.32\columnwidth}
    \includegraphics[width=\linewidth]{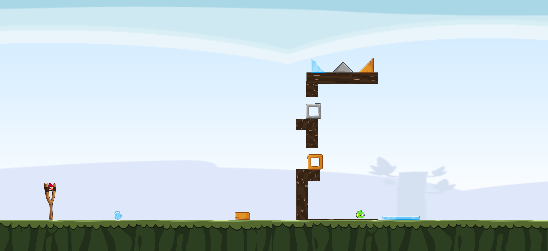}
    \caption{7.1}
  \end{subfigure}
  \begin{subfigure}[b]{0.32\columnwidth}
    \includegraphics[width=\linewidth]{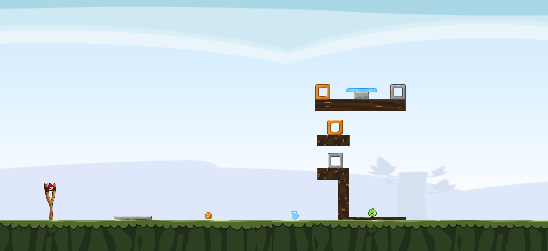}
    \caption{7.2}
  \end{subfigure}
    \begin{subfigure}[b]{0.32\columnwidth}
    \includegraphics[width=\linewidth]{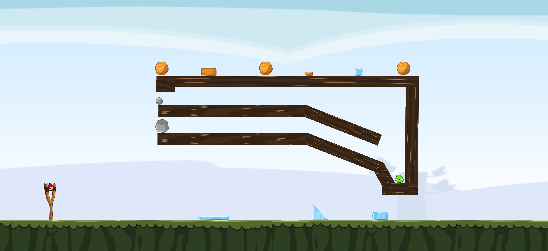}
    \caption{7.3}
  \end{subfigure}
  \begin{subfigure}[b]{0.32\columnwidth}
    \includegraphics[width=\linewidth]{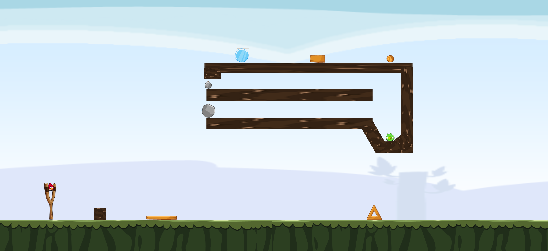}
    \caption{7.4}
  \end{subfigure}
  \begin{subfigure}[b]{0.32\columnwidth}
    \includegraphics[width=\linewidth]{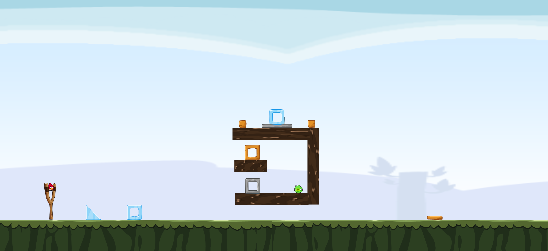}
    \caption{7.5}
  \end{subfigure}
\caption{Task templates from the relative weight scenario. The pig in these templates can be destroyed by shooting the bird at the block with the correct weight such that the block moves and collides with the pig.}
\label{event_7}
\end{figure}

\begin{figure}[H]
  \captionsetup[subfigure]{labelformat=empty}
  \centering
    \begin{subfigure}[b]{0.32\columnwidth}
    \includegraphics[width=\linewidth]{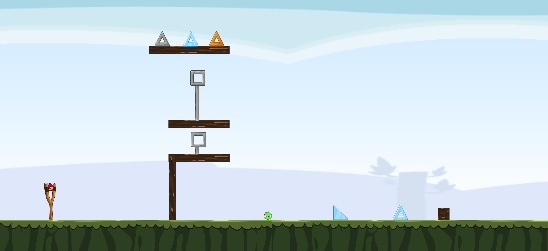}
    \caption{8.1}
  \end{subfigure}
  \begin{subfigure}[b]{0.32\columnwidth}
    \includegraphics[width=\linewidth]{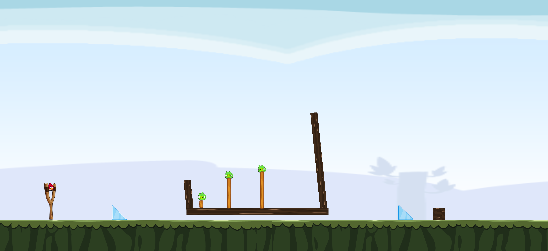}
    \caption{8.2}
  \end{subfigure}
  \begin{subfigure}[b]{0.32\columnwidth}
    \includegraphics[width=\linewidth]{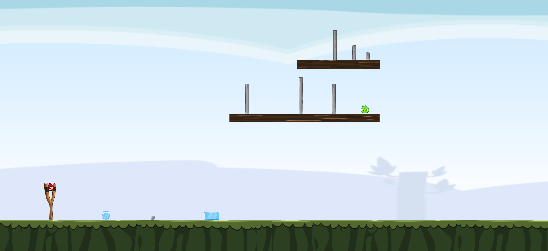}
    \caption{8.3}
  \end{subfigure}
  \begin{subfigure}[b]{0.32\columnwidth}
    \includegraphics[width=\linewidth]{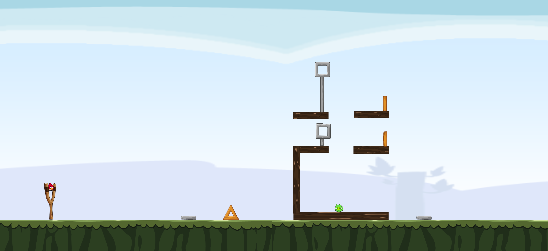}
    \caption{8.4}
  \end{subfigure}
\caption{Task templates from the relative height scenario. The pig in these templates can be destroyed by shooting the bird at the block with the correct height or at the block located at correct height such that the blocks move and collide with the pig.}
\label{event_8}
\end{figure}

\begin{figure}[H]
  \captionsetup[subfigure]{labelformat=empty}
  \centering
  \begin{subfigure}[b]{0.32\columnwidth}
    \includegraphics[width=\linewidth]{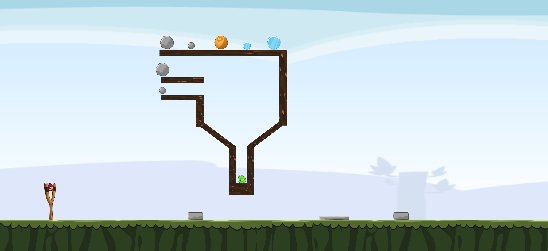}
    \caption{9.1}
  \end{subfigure}
  \begin{subfigure}[b]{0.32\columnwidth}
    \includegraphics[width=\linewidth]{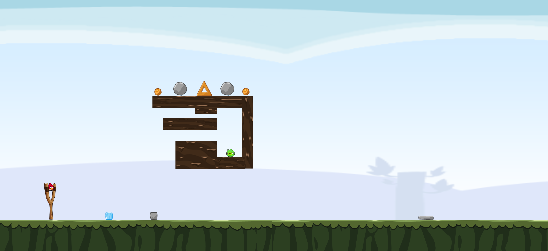}
    \caption{9.2}
  \end{subfigure}
  \begin{subfigure}[b]{0.32\columnwidth}
    \includegraphics[width=\linewidth]{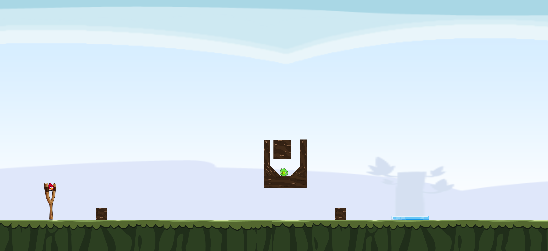}
    \caption{9.3}
  \end{subfigure}
  \begin{subfigure}[b]{0.32\columnwidth}
    \includegraphics[width=\linewidth]{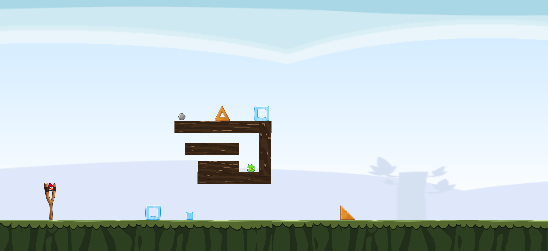}
    \caption{9.4}
  \end{subfigure}
\caption{Task templates from the relative width scenario. The pig in these templates can be destroyed by shooting the bird at the block with the correct width or using the bird itself to squeeze through the opening that it can fit and collide with the pig.}
\label{event_9}
\end{figure}

\begin{figure}[H]
  \captionsetup[subfigure]{labelformat=empty}
  \centering
   \begin{subfigure}[b]{0.32\columnwidth}
    \includegraphics[width=\linewidth]{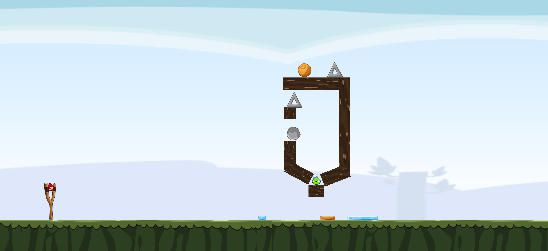}
    \caption{10.1}
  \end{subfigure}
  \begin{subfigure}[b]{0.32\columnwidth}
    \includegraphics[width=\linewidth]{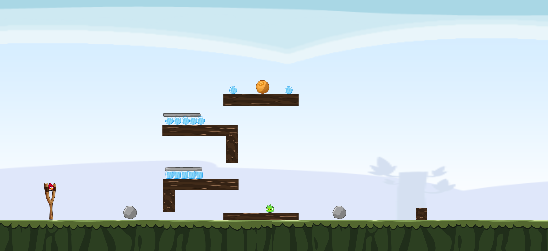}
    \caption{10.2}
  \end{subfigure}
  \begin{subfigure}[b]{0.32\columnwidth}
    \includegraphics[width=\linewidth]{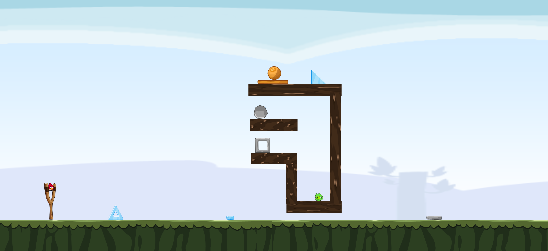}
    \caption{10.3}
  \end{subfigure}
  \begin{subfigure}[b]{0.32\columnwidth}
    \includegraphics[width=\linewidth]{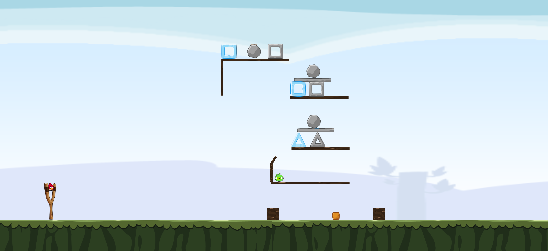}
    \caption{10.4}
  \end{subfigure}
\caption{Task templates from the shape difference scenario. The pig in these templates can be destroyed by shooting the bird considering the consequences of the shot due to the effect of the differences in the objects' shapes.}
\label{event_10}
\end{figure}

\begin{figure}[H]
  \captionsetup[subfigure]{labelformat=empty}
  \centering
  \begin{subfigure}[b]{0.32\columnwidth}
    \includegraphics[width=\linewidth]{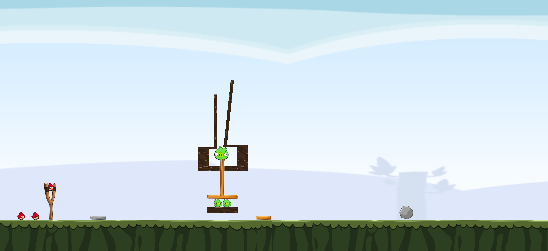}
    \caption{11.1}
  \end{subfigure}
  \begin{subfigure}[b]{0.32\columnwidth}
    \includegraphics[width=\linewidth]{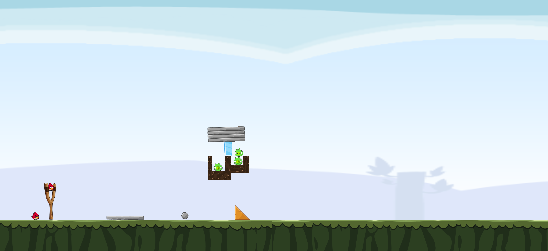}
    \caption{11.2}
  \end{subfigure}
  \begin{subfigure}[b]{0.32\columnwidth}
    \includegraphics[width=\linewidth]{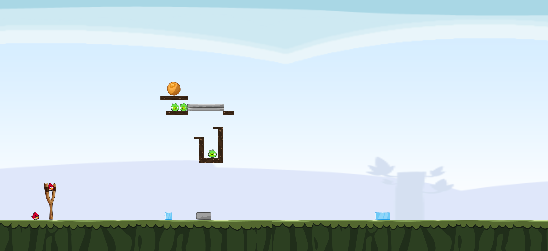}
    \caption{11.3}
  \end{subfigure}
  \begin{subfigure}[b]{0.32\columnwidth}
    \includegraphics[width=\linewidth]{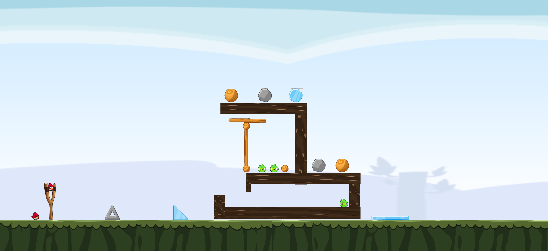}
    \caption{11.4}
  \end{subfigure}
  \begin{subfigure}[b]{0.32\columnwidth}
    \includegraphics[width=\linewidth]{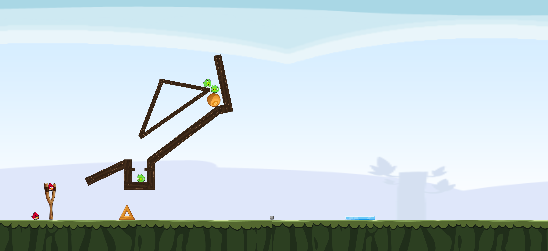}
    \caption{11.5}
  \end{subfigure}
\caption{Task templates from the non-greedy actions scenario. A greedy action in Angry Birds can be considered as the action that kills a higher number of pigs. In these templates, the non-greedy action needs to be done first since the level becomes unsolvable if the greedy action is done first.}
\label{event_11}
\end{figure}

\begin{figure}[H]
  \captionsetup[subfigure]{labelformat=empty}
  \centering
  \begin{subfigure}[b]{0.32\columnwidth}
    \includegraphics[width=\linewidth]{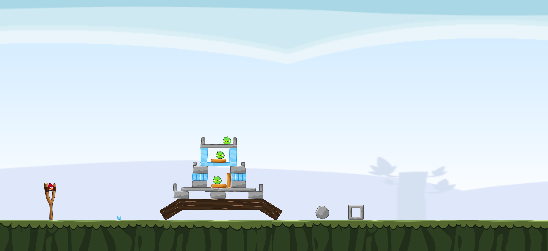}
    \caption{12.1}
  \end{subfigure}
  \begin{subfigure}[b]{0.32\columnwidth}
    \includegraphics[width=\linewidth]{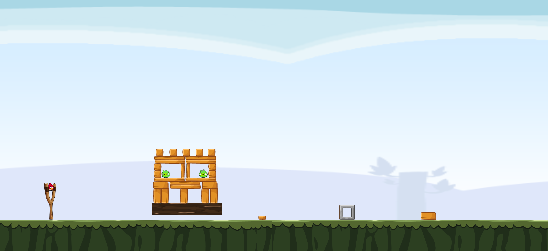}
    \caption{12.2}
  \end{subfigure}
  \begin{subfigure}[b]{0.32\columnwidth}
    \includegraphics[width=\linewidth]{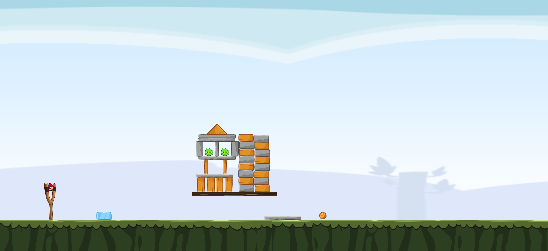}
    \caption{12.3}
  \end{subfigure}
  \begin{subfigure}[b]{0.32\columnwidth}
    \includegraphics[width=\linewidth]{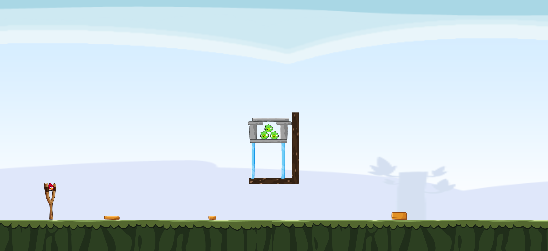}
    \caption{12.4}
  \end{subfigure}
  \begin{subfigure}[b]{0.32\columnwidth}
    \includegraphics[width=\linewidth]{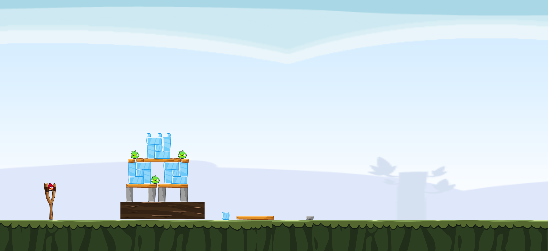}
    \caption{12.5}
  \end{subfigure}
  \begin{subfigure}[b]{0.32\columnwidth}
    \includegraphics[width=\linewidth]{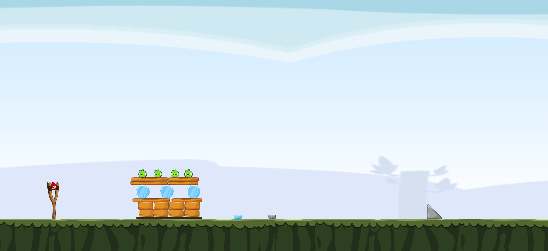}
    \caption{12.6}
  \end{subfigure}
\caption{Task templates from the structural analysis scenario. The pigs in these templates can be destroyed by shooting the bird at the correct position of the physical structure to break its stability and collapse.}
\label{event_12}
\end{figure}

\begin{figure}[H]
  \captionsetup[subfigure]{labelformat=empty}
  \centering
  \begin{subfigure}[b]{0.32\columnwidth}
    \includegraphics[width=\linewidth]{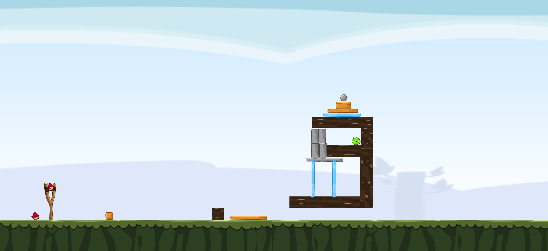}
    \caption{13.1}
  \end{subfigure}
  \begin{subfigure}[b]{0.32\columnwidth}
    \includegraphics[width=\linewidth]{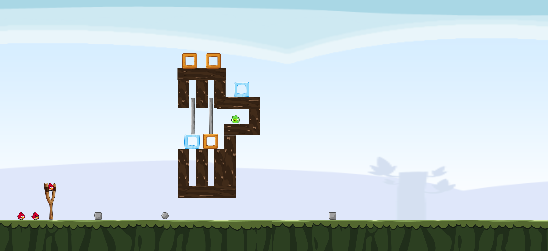}
    \caption{13.2}
  \end{subfigure}
  \begin{subfigure}[b]{0.32\columnwidth}
    \includegraphics[width=\linewidth]{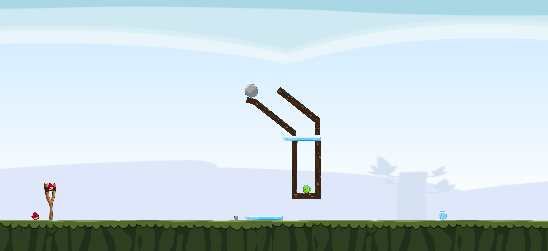}
    \caption{13.3}
  \end{subfigure}
  \begin{subfigure}[b]{0.32\columnwidth}
    \includegraphics[width=\linewidth]{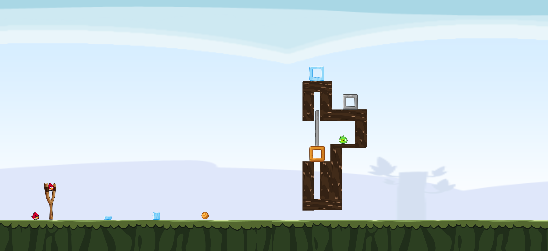}
    \caption{13.4}
  \end{subfigure}
  \begin{subfigure}[b]{0.32\columnwidth}
    \includegraphics[width=\linewidth]{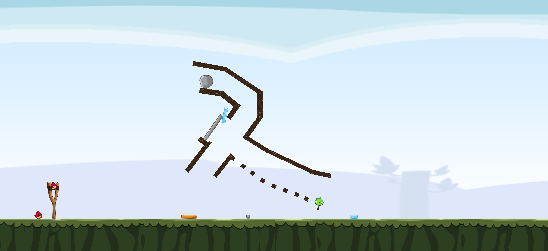}
    \caption{13.5}
  \end{subfigure}
\caption{Task templates from the clearing paths scenario. In these templates, the path to reach the pig needs to be cleared/created first and then destroy the pig.}
\label{event_13}
\end{figure}

\begin{figure}[H]
  \captionsetup[subfigure]{labelformat=empty}
  \centering
  \begin{subfigure}[b]{0.32\columnwidth}
    \includegraphics[width=\linewidth]{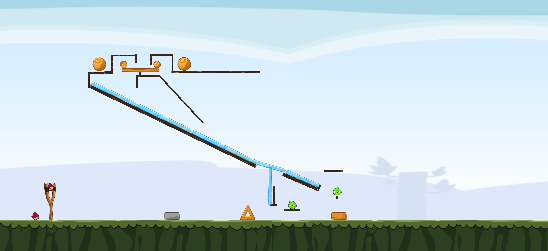}
    \caption{14.1}
  \end{subfigure}
  \begin{subfigure}[b]{0.32\columnwidth}
    \includegraphics[width=\linewidth]{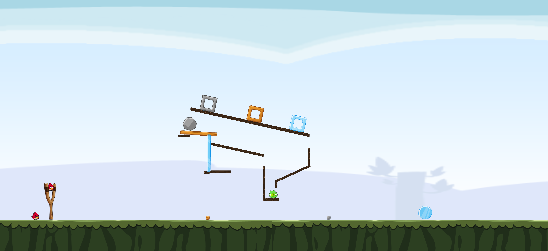}
    \caption{14.2}
  \end{subfigure}
\caption{Task templates from the adequate timing scenario. In these templates, after doing the first bird shot, the second bird is needed to shoot within a time frame such that all the pigs can be destroyed.}
\label{event_14}
\end{figure}

\begin{figure}[H]
  \captionsetup[subfigure]{labelformat=empty}
  \centering
    \begin{subfigure}[b]{0.32\columnwidth}
    \includegraphics[width=\linewidth]{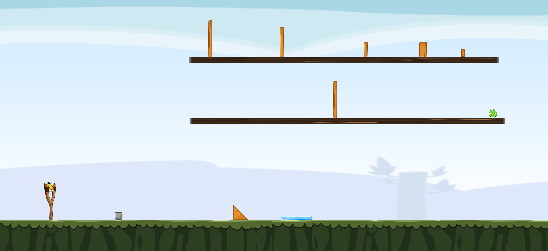}
    \caption{15.1}
  \end{subfigure}
  \begin{subfigure}[b]{0.32\columnwidth}
    \includegraphics[width=\linewidth]{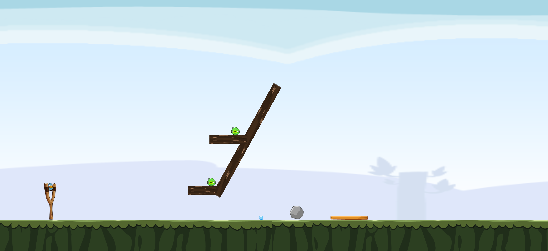}
    \caption{15.2}
  \end{subfigure}
  \begin{subfigure}[b]{0.32\columnwidth}
    \includegraphics[width=\linewidth]{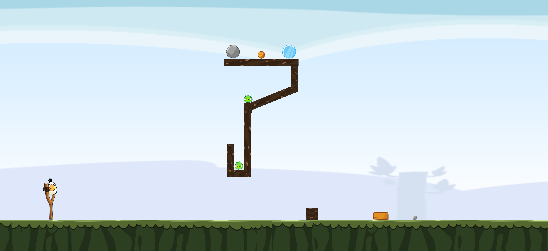}
    \caption{15.3}
  \end{subfigure}
  \begin{subfigure}[b]{0.32\columnwidth}
    \includegraphics[width=\linewidth]{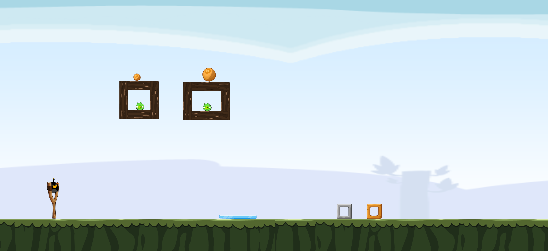}
    \caption{15.4}
  \end{subfigure}
  \begin{subfigure}[b]{0.32\columnwidth}
    \includegraphics[width=\linewidth]{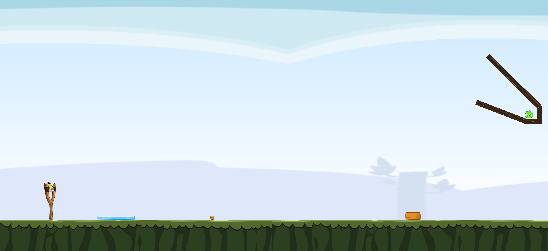}
    \caption{15.5}
  \end{subfigure}
  \begin{subfigure}[b]{0.32\columnwidth}
    \includegraphics[width=\linewidth]{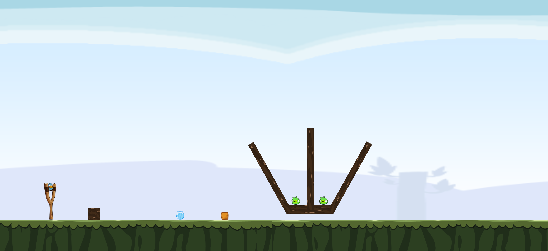}
    \caption{15.6}
  \end{subfigure}
  \begin{subfigure}[b]{0.32\columnwidth}
    \includegraphics[width=\linewidth]{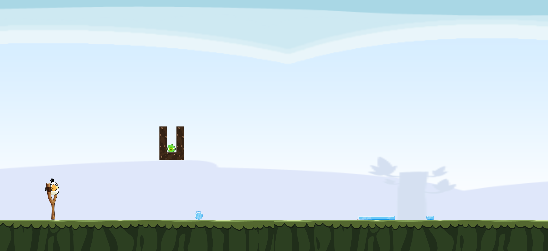}
    \caption{15.7}
  \end{subfigure}
  \begin{subfigure}[b]{0.32\columnwidth}
    \includegraphics[width=\linewidth]{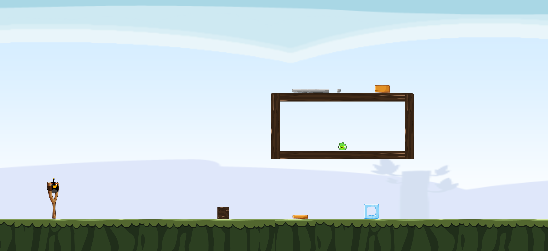}
    \caption{15.8}
  \end{subfigure}
\caption{Task templates from the manoeuvring scenario. The pigs in these templates can be destroyed by correctly manoeuvring the birds by activating their powers in flight.}
\label{event_15}
\end{figure}

\begin{figure}[H]
  \captionsetup[subfigure]{labelformat=empty}
  \centering
  \begin{subfigure}[b]{0.24\columnwidth}
    \includegraphics[width=\linewidth]{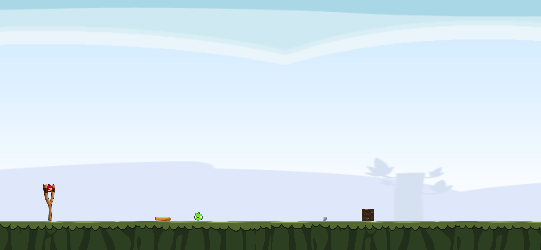}
  \end{subfigure}
\begin{subfigure}[b]{0.24\columnwidth}
    \includegraphics[width=\linewidth]{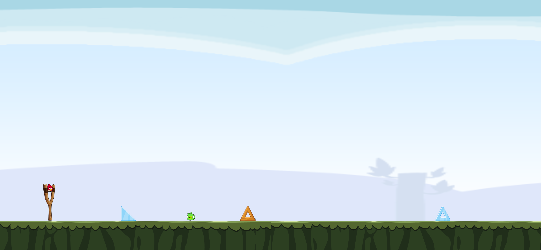}
  \end{subfigure}
  \begin{subfigure}[b]{0.24\columnwidth}
    \includegraphics[width=\linewidth]{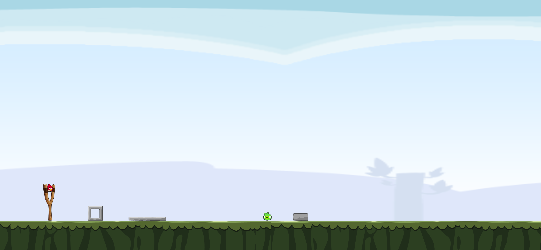}
  \end{subfigure}
  \begin{subfigure}[b]{0.24\columnwidth}
    \includegraphics[width=\linewidth]{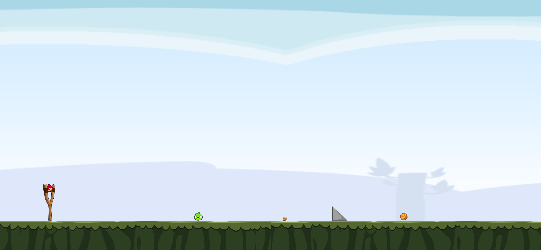}
  \end{subfigure}
  \begin{subfigure}[b]{0.24\columnwidth}
    \includegraphics[width=\linewidth]{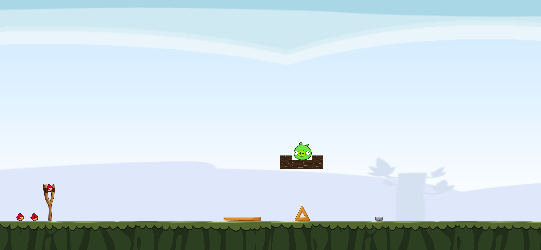}
  \end{subfigure}
  \begin{subfigure}[b]{0.24\columnwidth}
    \includegraphics[width=\linewidth]{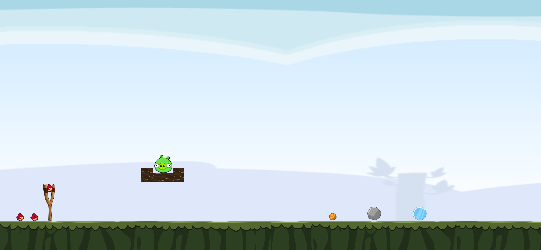}
  \end{subfigure}
  \begin{subfigure}[b]{0.24\columnwidth}
    \includegraphics[width=\linewidth]{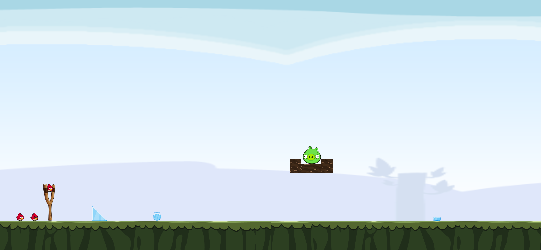}
  \end{subfigure}
  \begin{subfigure}[b]{0.24\columnwidth}
    \includegraphics[width=\linewidth]{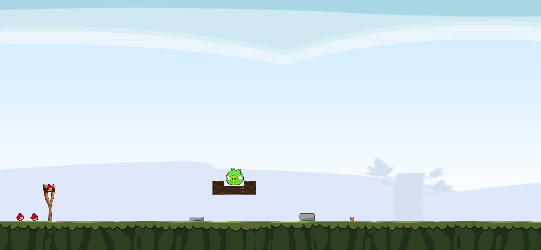}
  \end{subfigure}
  \begin{subfigure}[b]{0.24\columnwidth}
    \includegraphics[width=\linewidth]{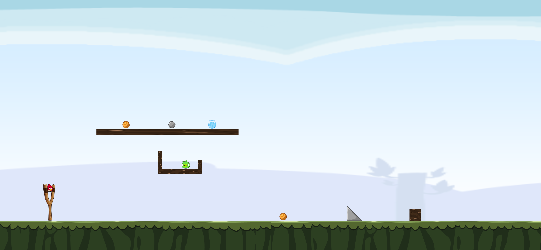}
  \end{subfigure}
  \begin{subfigure}[b]{0.24\columnwidth}
    \includegraphics[width=\linewidth]{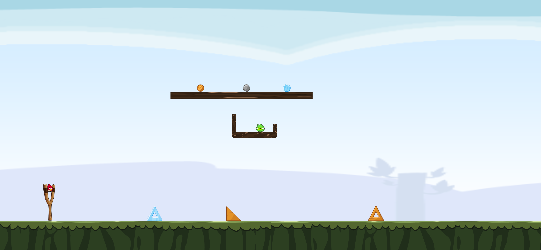}
  \end{subfigure}
  \begin{subfigure}[b]{0.24\columnwidth}
    \includegraphics[width=\linewidth]{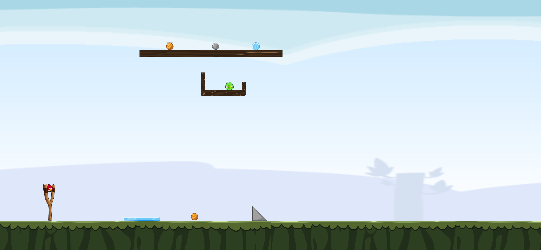}
  \end{subfigure}
    \begin{subfigure}[b]{0.24\columnwidth}
    \includegraphics[width=\linewidth]{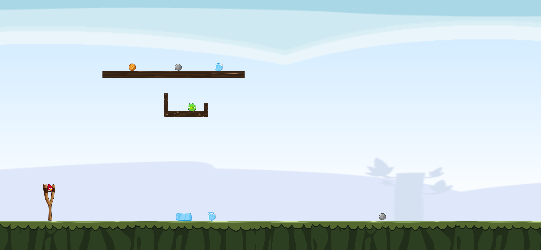}
  \end{subfigure}
  \begin{subfigure}[b]{0.24\columnwidth}
    \includegraphics[width=\linewidth]{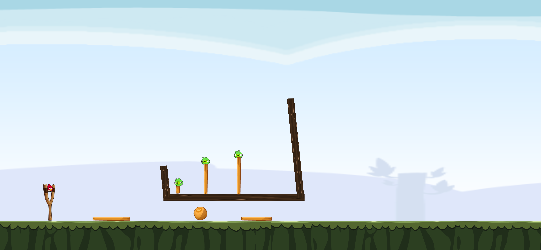}
  \end{subfigure}
  \begin{subfigure}[b]{0.24\columnwidth}
    \includegraphics[width=\linewidth]{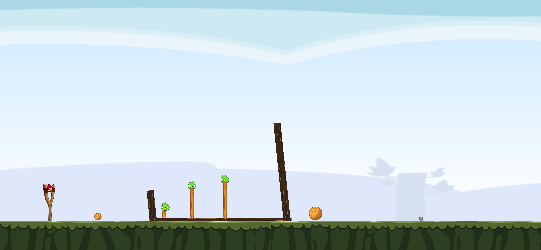}
  \end{subfigure}
  \begin{subfigure}[b]{0.24\columnwidth}
    \includegraphics[width=\linewidth]{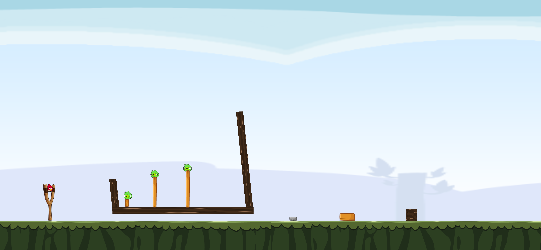}
  \end{subfigure}
  \begin{subfigure}[b]{0.24\columnwidth}
    \includegraphics[width=\linewidth]{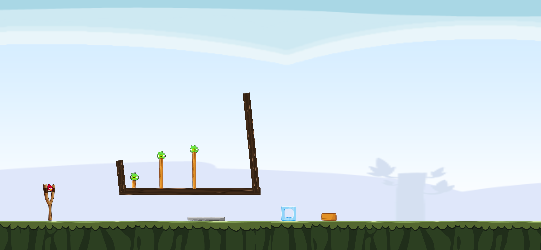}
  \end{subfigure}
    \begin{subfigure}[b]{0.24\columnwidth}
    \includegraphics[width=\linewidth]{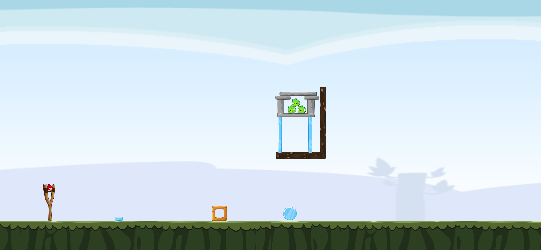}
  \end{subfigure}
  \begin{subfigure}[b]{0.24\columnwidth}
    \includegraphics[width=\linewidth]{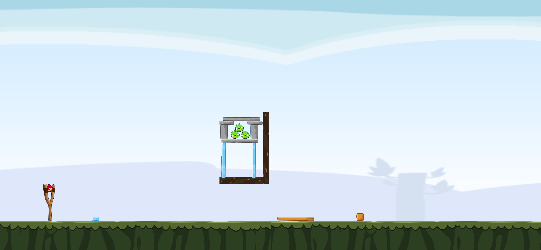}
  \end{subfigure}
  \begin{subfigure}[b]{0.24\columnwidth}
    \includegraphics[width=\linewidth]{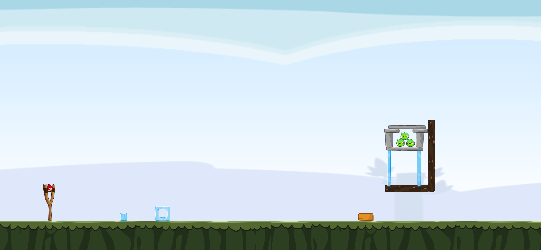}
  \end{subfigure}
  \begin{subfigure}[b]{0.24\columnwidth}
    \includegraphics[width=\linewidth]{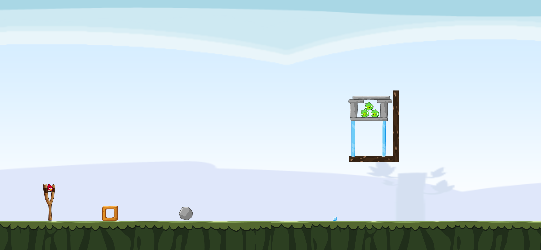}
  \end{subfigure}
      \begin{subfigure}[b]{0.24\columnwidth}
    \includegraphics[width=\linewidth]{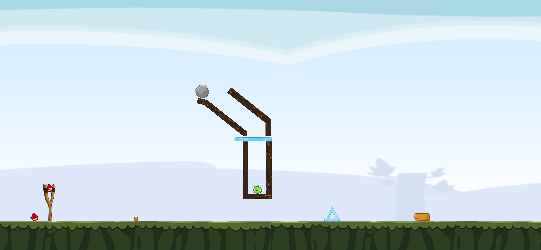}
  \end{subfigure}
  \begin{subfigure}[b]{0.24\columnwidth}
    \includegraphics[width=\linewidth]{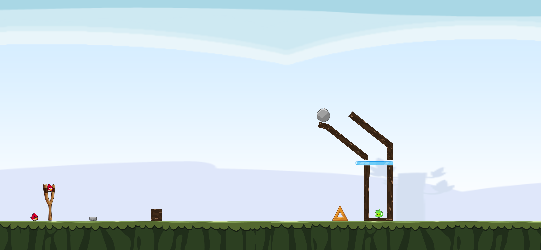}
  \end{subfigure}
  \begin{subfigure}[b]{0.24\columnwidth}
    \includegraphics[width=\linewidth]{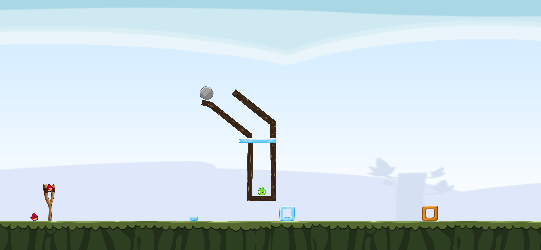}
  \end{subfigure}
  \begin{subfigure}[b]{0.24\columnwidth}
    \includegraphics[width=\linewidth]{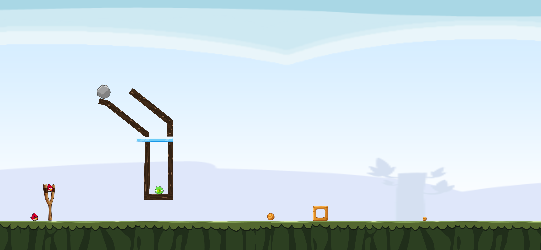}
  \end{subfigure}
\caption{Each row shows four example tasks generated from the same task template. The positions of the objects and the distraction objects vary within the tasks of the same template, therefore each task has its own solution; however, all tasks of the same template can be solved by the same physical rule.}
\label{task_variations_in_hiphy}
\end{figure}

Phy-Q contains 75 task templates for 15 physical scenarios. The figures \ref{event_1} to \ref{event_15} show the task templates of the 15 scenarios: single force, multiple forces, rolling, falling, sliding, bouncing, relative weight, relative height, relative width, shape difference, non-greedy actions, structural analysis, clearing paths, adequate timing, and manoeuvring respectively. Figure \ref{task_variations_in_hiphy} shows the tasks generated from six example task templates.

\subsection{Differences Between the Task Templates in Phy-Q}
\begin{table}[H]
\small
  \caption{The differences between the task templates of the scenarios from 1 to 9 are presented using the high-level solution descriptions of the task templates.}
  \label{task_difference_1}
  \centering
  \begin{tabularx}{\textwidth}{lX}
    \toprule
    Scenario & Description\\
    \midrule
    1. Single & 1.1 Applying a single force using the lower trajectory of the bird.\\
    force & 1.2 Applying a single force using the higher trajectory of the bird.\\
    & 1.3 Applying a single adequate force using a specific stretch of the slingshot.\\
    & 1.4 Applying a single force to the pig on the ground by using either lower or higher trajectory of the bird.\\
    & 1.5 Applying a single force to the pig on the platform by using either lower or higher trajectory of the bird.\\
    \hline
    2. Multiple & 2.1 Applying multiple forces using the lower trajectory of the bird.\\
    forces & 2.2 Applying multiple forces using the higher trajectory of the bird.\\
    & 2.3 Applying multiple adequate forces using a specific stretch of the slingshot.\\
    & 2.4 Applying multiple forces to the pig on the ground by using either lower or higher trajectory of the bird.\\
    & 2.5 Applying multiple forces to the pig on the platform by using either lower or higher trajectory of the bird.\\
    \hline
    3. Rolling & 3.1 Rolling the bird down a ramp into the directly-unreachable pig.\\
    & 3.2 Rolling the circular object on a flat surface into the path-blocked pig.\\
    & 3.3 Rolling the circular object down a ramp into the path-blocked pig.\\
    & 3.4 Rolling the bird on the flat surface into the directly-unreachable pig.\\
    & 3.5 Rolling the circular object into the directly-unreachable pig.\\
    & 3.6 Rolling the circular object down a ramp into the directly-unreachable pig.\\
    \hline
    4. Falling & 4.1 Falling the square block choosing the correct path onto the pig.\\
    & 4.2 Falling the circular block choosing the correct path on to the pig.\\
    & 4.3 Falling the two circular blocks onto the two pigs.\\
    & 4.4 Falling the circular block onto the pig.\\
    & 4.5 Falling the square block onto the pig.\\
    \hline
    5. Sliding & 5.1 Sliding the block on a flat surface into the unreachable pig.\\
    & 5.2 Sliding the block down a ramp into the path-blocked pig.\\
    & 5.3 Sliding the block on a flat surface into the path-blocked pig.\\
    & 5.4 Sliding the block on the ground into the unreachable pig.\\
    & 5.5 Sliding the block on the ground into the path-blocked pig.\\
    \hline
    6. Bouncing & 6.1 Bouncing the bird off a flat surface onto the pig in an inclined plane.\\
    & 6.2 Bouncing the bird upwards by smashing the bird on a lower surface to reach the pig. \\
    & 6.3 Bouncing the bird off an inclined surface onto the pig in the pouch.\\
    & 6.4 Bouncing the bird off a flat surface onto the pig on a flat surface.\\
    & 6.5 Bouncing the bird upwards by smashing the bird on a lower surface to reach the big pig.\\
    & 6.6 Bouncing the bird off an inclined surface onto the pig in an inclined plane. \\
    \hline
    7. Relative weight & 7.1 Falling the correct block (below block) to fall onto the pig considering the weight. \\
    & 7.2 Falling the correct block (above block) to fall onto the pig considering the weight.\\
    & 7.3 Rolling the small circular block with less weight on a flat surface and on an inclined surface to reach the pig. \\
    & 7.4 Rolling the small circular block with less weight on a flat surface to reach the pig.\\
    & 7.5 Sliding and falling the brown coloured block onto the pig considering the weight.\\ 
    \hline
    8. Relative height & 8.1 Toppling the square block on the taller block onto the pig. \\
    & 8.2 Bouncing the bird to hit the tallest block in order to create a domino effect.\\
    & 8.3 Shooting the tallest block to create a domino effect to reach the pig.\\
    & 8.4 Falling the square block on the short block onto the pig.\\
    \hline
    9. Relative width & 9.1 Falling the small circular block that can go through the slit.\\
    & 9.2 Shooting the bird using the lower trajectory at the lower opening which is wider to squeeze the bird through it.\\
    & 9.3 Shooting the bird using the higher trajectory at the wider opening to fall into it. \\
    & 9.4 Shooting the bird using the lower trajectory at the higher opening which is wider to squeeze the bird through it.\\
    \hline
  \end{tabularx}
\end{table}

\newpage

\begin{table}[h!]
\small
  \caption{The differences between the task templates of the scenarios from 10 to 15 are presented using the high-level solution descriptions of the task templates.}
  \label{task_difference_2}
  \centering
  \begin{tabularx}{\textwidth}{lX}
    \toprule
    Scenario & Description\\
    \midrule
    10. Shape difference & 10.1 Shooting at the triangular block instead of the circular block, so that the edges of the triangle can reach the pig.\\
    & 10.2 Shooting at the circular block supporter (instead of the square block supporter) which can make the rectangular block to roll and fall on the pig. \\
    & 10.3 Shooting at the circular block (instead of the square block) which can roll and fall on the pig. \\
    & 10.4 Shooting at the triangular block supporter (instead of the square block supporter) which enables the circular object to roll onto the pig. \\
    \hline
    11. Non-greedy actions & 11.1 Using the first two birds to destroy the larger pig (non-greedy action) and shooting the last bird to destroy the two pigs.\\
    & 11.2 Shooting the first bird to destroy the pig in the front (non-greedy action) before destroying the two pigs. \\
    & 11.3 Taking the non-greedy action first by shooting at the basket with one pig before shooting at the two pigs.\\
    & 11.4 Taking the non-greedy action first by targeting the tunnel with a single pig before shooting at the location with two pigs.\\
    & 11.5 Taking the non-greedy action first to destroy the pig inside the basket before shooting the two pigs on top. \\
    \hline
    12. Structural analysis & 12.1 - 12.6 Shooting at the correct position to break the stability and destroy the pigs based on the given structure.\\
    \hline
    13. Clearing paths & 13.1 Shooting the first bird to destroy the supporting blocks to clear the path to the pig.\\
    & 13.2 Shooting the first two birds to fall the path-blocking blocks into the slits.\\
    & 13.3 Shooting at the blocking-block to clear the path for the circular block to reach the pig. \\
    & 13.4 Shooting the first bird to fall the path-blocking block into the slit.\\
    & 13.5 Using the first bird to fall the block to create a path to reach the pig. \\
    \hline
    14. Adequate timing & 14.1 Shooting the second bird at the correct time after the two balls on top move down the ramp. \\
    & 14.2 Shooting the second bird at the correct time before the circular block blocks the path to the pig. \\
    \hline 
    15. Manoeuvring & 15.1 Tapping the yellow bird before reaching the wood block to reach the pig.\\
    & 15.2 Tapping the blue bird before reaching the pig on the top to make sure two child blue birds reach the two pigs. \\
    & 15.3 Tapping the white bird to ensure the egg reaches the lower pig and the bird reaches the upper pig. \\
    & 15.4 Tapping the black bird to cause an explosion that affects both pigs.\\
    & 15.5 Tapping the yellow bird to reach the unreachable pig. \\
    & 15.6 Tapping the blue bird to ensure child birds fall to the two separate baskets with pigs. \\
    & 15.7 Tapping the white bird to ensure the egg falls onto the pig. \\
    & 15.8 Tapping the black bird closer to the covered pig. \\
    \hline
  \end{tabularx}
\end{table}

\newpage

\newpage

\subsection{Solution Descriptions for Example Task Templates in Phy-Q}
Table \ref{exmaple_solutions} shows solution descriptions for 15 example task templates in Phy-Q representing one from each 15 physical scenarios.

\begin{table}[h!]
  \caption{Solution descriptions for 15 example task templates in Phy-Q.}
  \label{exmaple_solutions}
  \centering
  \begin{tabularx}{\textwidth}{lX}
    \toprule
    Task template & Solution description\\
    \midrule
    1.4 & Single force: A single force is needed to be applied to the pig to destroy it by a direct bird shot.\\
    2.5 & Multiple forces: Multiple forces are needed to be applied to destroy the pig by multiple bird shots.\\
    3.6 & Rolling: The circular object is needed to be rolled onto the pig, which is unreachable for the bird from the slingshot, causing the pig to be destroyed.\\
    4.4 & Falling: The circular object is needed to be fallen onto the pig causing the pig to be destroyed.\\
    5.1 & Sliding: The square object is needed to be slid to hit the pig, which is unreachable for the bird from the slingshot, causing the pig to be destroyed.\\
    6.1 & Bouncing: The bird is needed to be bounced off the platform (dark-brown object) to hit and destroy the pig.\\
    7.3 & Relative weight: The small circular block is lighter than the big circular block. Out of the two blocks, the small circular block can only be rolled to reach the pig and destroy it.\\
    8.4 & Relative height: The square block on top of the taller rectangular block will not fall through the gap due to the height of the rectangular block. Hence the square block on top of the shorter rectangular block needs to be toppled to fall through the gap and destroy the pig.\\
    9.4 & Relative width: The bird cannot go through the lower entrance which has a narrow opening. Hence the bird is needed to be shot to the upper entrance to reach the pig and destroy it.\\
    10.4 & Shape difference: The circular block on two triangle blocks can be rolled down by breaking a one triangle block and the circular block on two square blocks cannot be rolled down by breaking a one square block. Hence, the triangle block needs to be destroyed to make the circular block roll and fall onto the pig causing the pig to be destroyed.\\
    11.5 & Non-greedy actions: A greedy action tries to destroy the highest number of pigs in a single bird shot. If the two pigs resting on the circular block are destroyed, then the circular block will roll down and block the entrance to reach the below pig. Hence, the below pig is needed to be destroyed first and then the upper two pigs.\\
    12.3 & Structural analysis: The bird is needed to be shot at the weak point of the structure to break the stability and destroy the pigs. Shooting elsewhere does not destroy the two pigs with a single bird shot.\\
    13.5 & Clearing paths: First, the rectangle block is needed to be positioned correctly to open the path for the circular block to reach the pig. Then the circular block is needed to be rolled to destroy the pig.\\
    14.4 & Adequate timing: First, the two circular objects are needed to be rolled to the ramp. Then, after the first circle passes the prop and before the second circle reaches the prop, the prop needs to be destroyed to make the second circle fall onto the lower pig.\\
    15.6 & Manoeuvring: The blue bird splits into three other birds when it is tapped in the flight. The blue bird is needed to be tapped at the correct position to manoeuvre the birds to reach the two separated pigs.\\
    \bottomrule
  \end{tabularx}
\end{table}

\newpage

\section{Local Genaralization Results}

\begin{figure}[h!]
  \centering
  \includegraphics[width=1.0\linewidth]{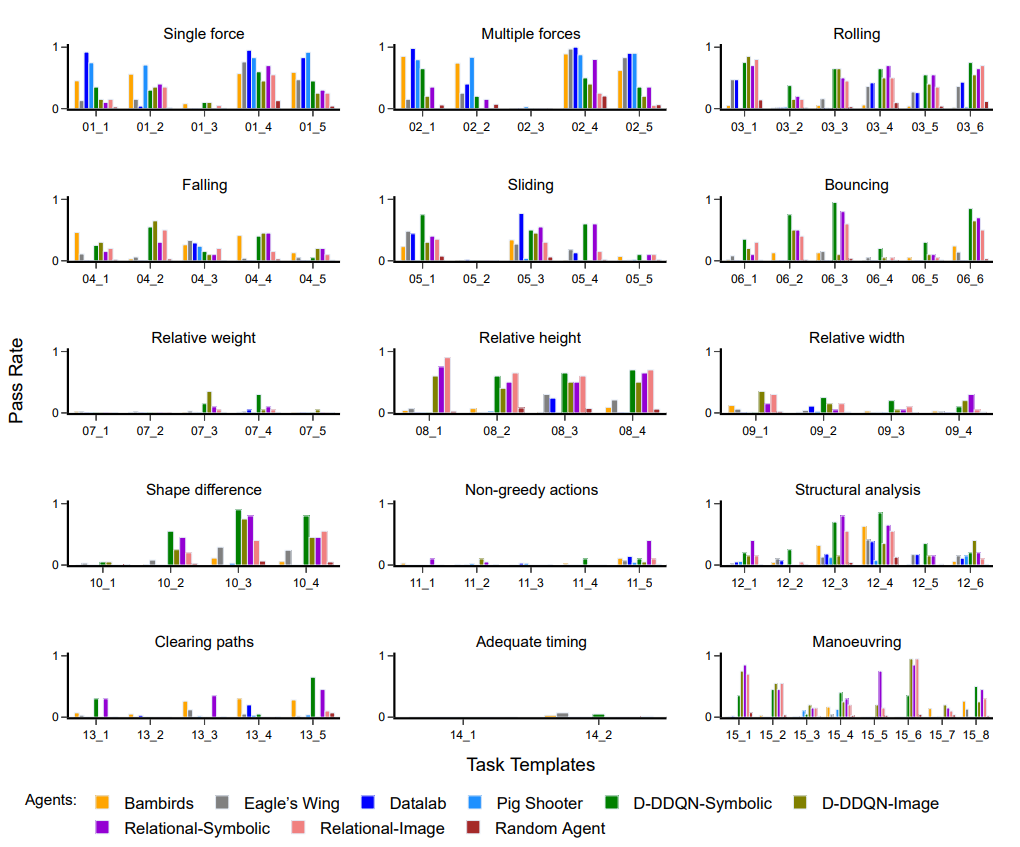}
  \caption{Passing rates of the agents for the \textit{within template evaluation}. The results are presented for the 75 task templates by separating them into the 15 physical scenarios (1. single force, 2. multiple forces, 3. rolling, 4. falling, 5. sliding, 6. bouncing, 7. relative weight, 8. relative height, 9. relative width, 10. shape difference, 11. non-greedy actions, 12. structural analysis, 13. clearing paths, 14. adequate timing, and 15. manoeuvring). The x-axis is the index of the task template (x\_y represents the y\textsuperscript{th} task template of the x\textsuperscript{th} physical scenario) and y-axis is the passing rate.}
  \label{fig_local_generalization}
\end{figure}

Figure \ref{fig_local_generalization} shows the passing rate of the nine baseline agents for the 75 task templates for the \textit{within template evaluation}. The within template evaluation measures the local generalization ability of an agent. For learning agents, we take the first 80 tasks from each task template as the training set and the last 20 tasks to evaluate the local generalization ability of the agent. As there is no training phase for heuristic agents, the heuristic agents are evaluated on the 20 test task templates. Due to the randomness in the heuristic agents, we give 5 attempts in each task and take the average pass rate of the task. Presented in Figure \ref{fig_local_generalization} is the average pass rate of all test tasks. Overall, it can be seen that learning agents perform better in the within template evaluation. 

\newpage

\section{Broad generalization Training and Testing Split}
Within scenario evaluation measures the broad generalization of an agent. Agent trains on a subset of task templates within a specific scenario and is evaluated on the rest of the task templates in the same scenario. Table \ref{train-test-split-events} shows the division of task templates for training and testing in the \textit{within scenario evaluation} for the 15 physical scenarios.

\begin{table}[h!]
\small
  \caption{The train-test split of the task templates for the \textit{within scenario evaluation}}
  \label{train-test-split-events}
  \centering
  \begin{tabular}{lll}
    \toprule
    Scenario & Training templates & Testing templates \\
    \midrule
    1. Single force  & 1.1, 1.2, 1.3 & 1.4, 1.5\\ 
    2. Multiple forces & 2.1, 2.2, 2.3 & 2.4, 2.5 \\
    3. Rolling  & 3.1, 3.2, 3.3 & 3.4, 3.5, 3.6 \\ 
    4. Falling & 4.1, 4.2, 4.3 & 4.4, 4.5 \\
    5. Sliding & 5.1, 5.2, 5.3 & 5.4, 5.5 \\
    6. Bouncing & 6.1, 6.2, 6.3 & 6.4, 6.5, 6.6 \\
    7. Relative weight & 7.1, 7.2, 7.3 & 7.4, 7.5\\
    8. Relative height & 8.1, 8.2 & 8.3, 8.4 \\
    9. Relative width & 9.1, 9.2 & 9.3, 9.4 \\
    10. Shape difference & 10.1, 10.2 & 10.3, 10.4 \\
    11. Non-greedy actions & 11.1, 11.2, 11.3 & 11.4, 11.5 \\
    12. Structural analysis & 12.1, 12.2, 12.3 & 12.4, 12.5, 12.6 \\
    13. Clearing paths & 13.1, 13.2, 13.3 & 13.4, 13.5 \\
    14. Adequate timing & 14.1 & 14.2 \\
    15. Manoeuvring & 15.1, 15.2, 15.3, 15.4 & 15.5, 15.6, 15.7, 15.8 \\
    \bottomrule
  \end{tabular}
\end{table}

\newpage

\section{Human Player Evaluation}
\begin{figure}[h!]
  \centering
  \includegraphics[scale=0.48]{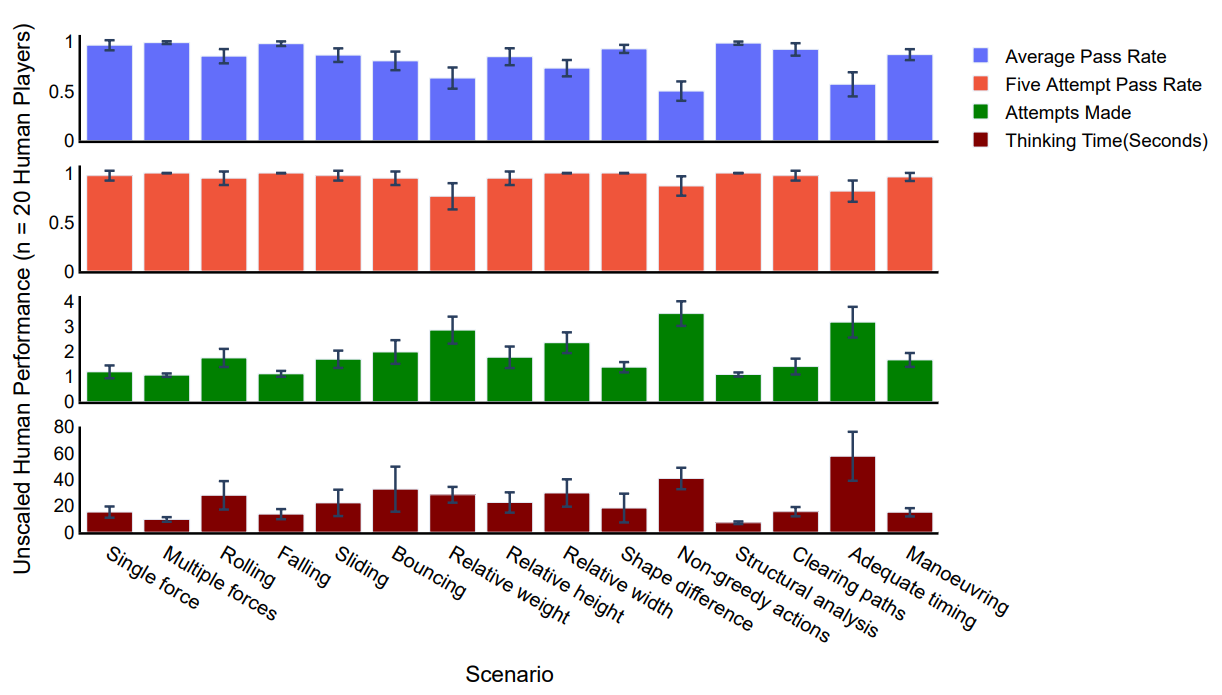}
  \caption{The average pass rate, five attempt pass rate,  maximum number of attempts made, and the total thinking time taken by human participants (in seconds) for the 15 physical scenarios. Error bars represent the 95\% confidence interval calculated using 20 human participants.}
  \label{fig_human_charts}
\end{figure}

\begin{figure}[ht]
  \centering
  \includegraphics[scale=0.58]{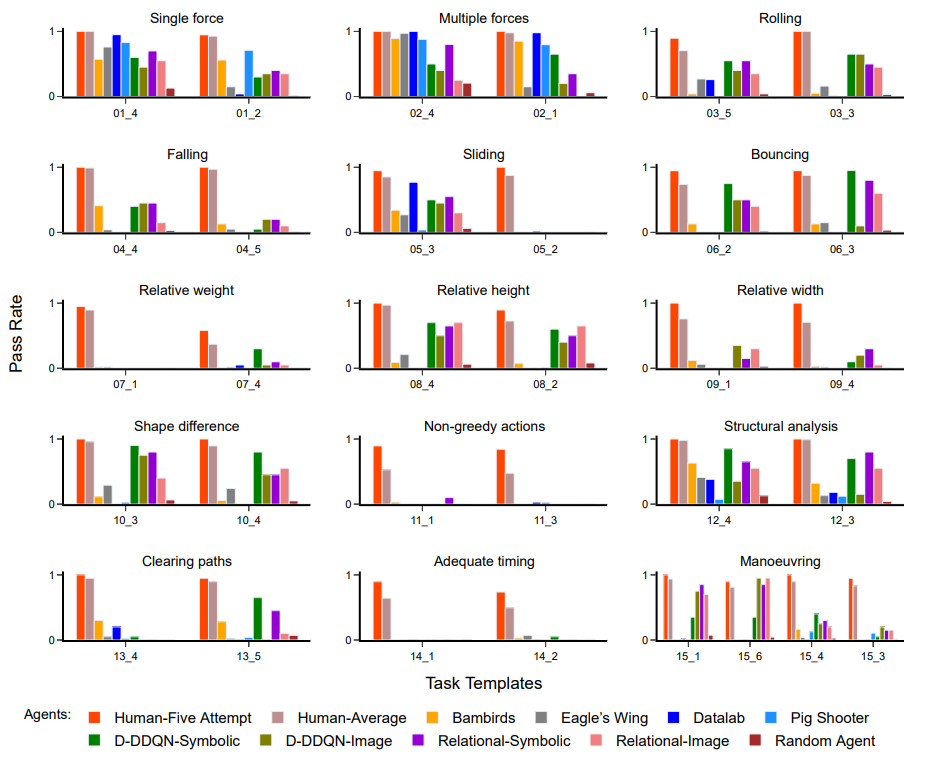}
  \caption{Performance of humans, the performance of heuristic agents, and local generalization of learning agents on the sample of task templates used for the human evaluation. It can be seen that overall human performance is far ahead of agents.}
  \label{fig_human_chart_all_templates}
\end{figure}

Figure \ref{fig_human_charts} presents the average pass rate, the pass rate the player achieved within five attempts, the maximum number of attempts made, and the total thinking time taken (in seconds) by human participants for the 15 capabilities. Figure \ref{fig_human_chart_all_templates} presents the performance of human participants, the performance of heuristic agents, and the local generalization of learning agents on the exact task templates used for the human player evaluation to make a direct comparison of agents with human performance. 
We used the average pass rates of the humans (e.g., if the player passes in the first attempt, the pass rate is 100\% and if the player passes in the fifth attempt the pass rate is 20\%) to have a fair comparison with the agents. We illustrate the five attempt pass rate to show that almost all the human players could pass the tasks within five attempts. 

\newpage

\section{2021 AIBIRDS Competition Game Levels Used for the Evaluation}

\begin{figure}[H]
  \captionsetup[subfigure]{labelformat=empty}
  \centering
  \begin{subfigure}[b]{0.32\columnwidth}
    \includegraphics[width=\linewidth]{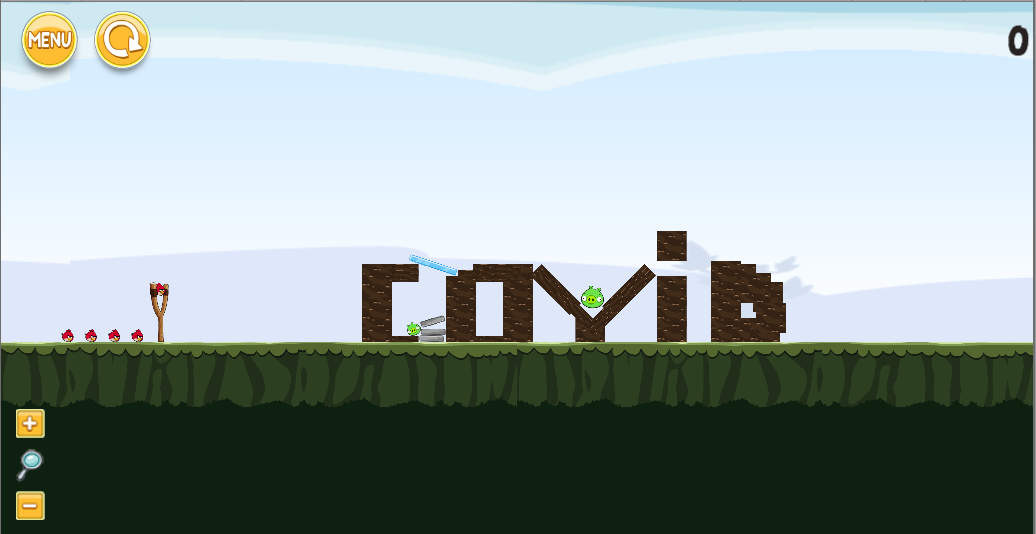}
    \caption{1}
  \end{subfigure}
  \begin{subfigure}[b]{0.32\columnwidth}
    \includegraphics[width=\linewidth]{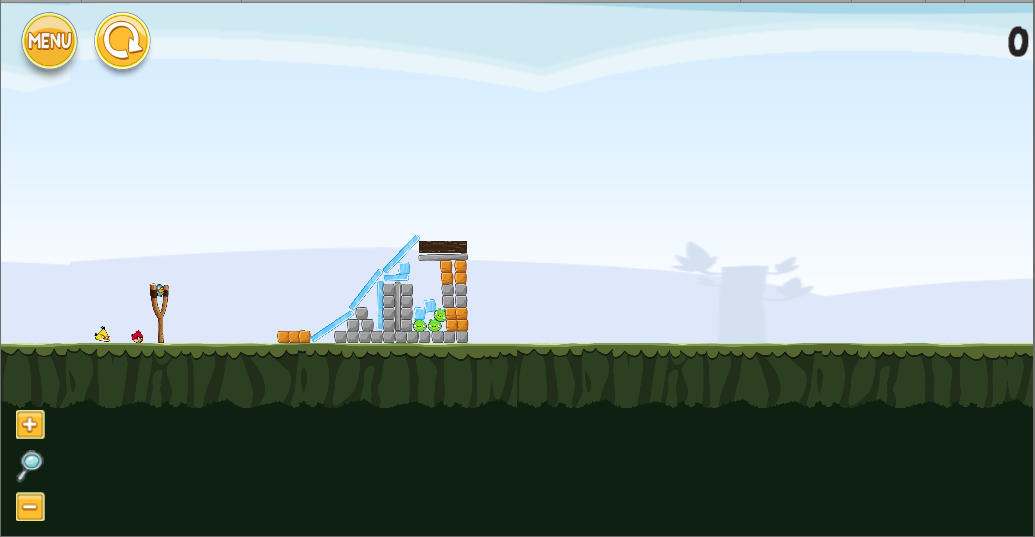}
    \caption{2}
  \end{subfigure}
  \begin{subfigure}[b]{0.32\columnwidth}
    \includegraphics[width=\linewidth]{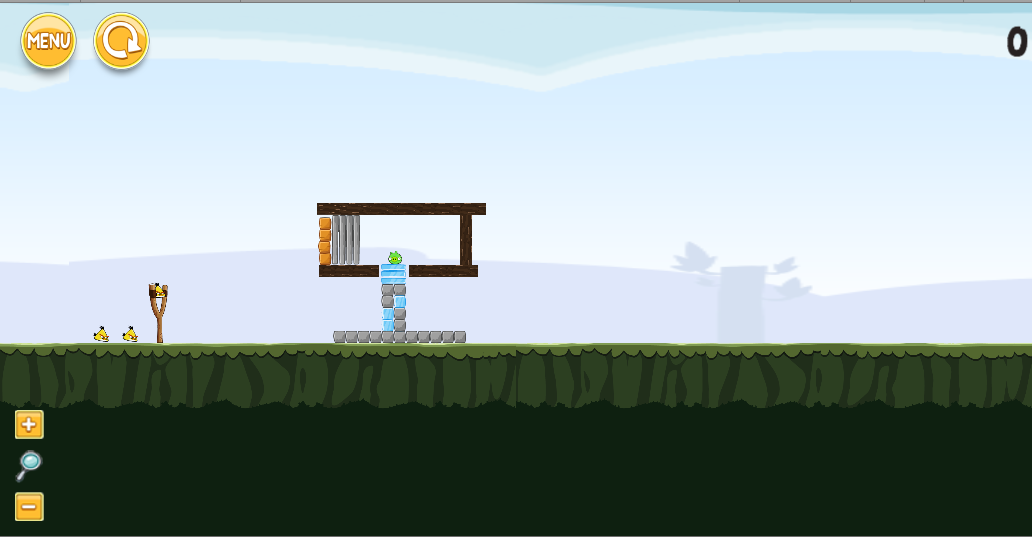}
    \caption{3}
  \end{subfigure}
  \begin{subfigure}[b]{0.32\columnwidth}
    \includegraphics[width=\linewidth]{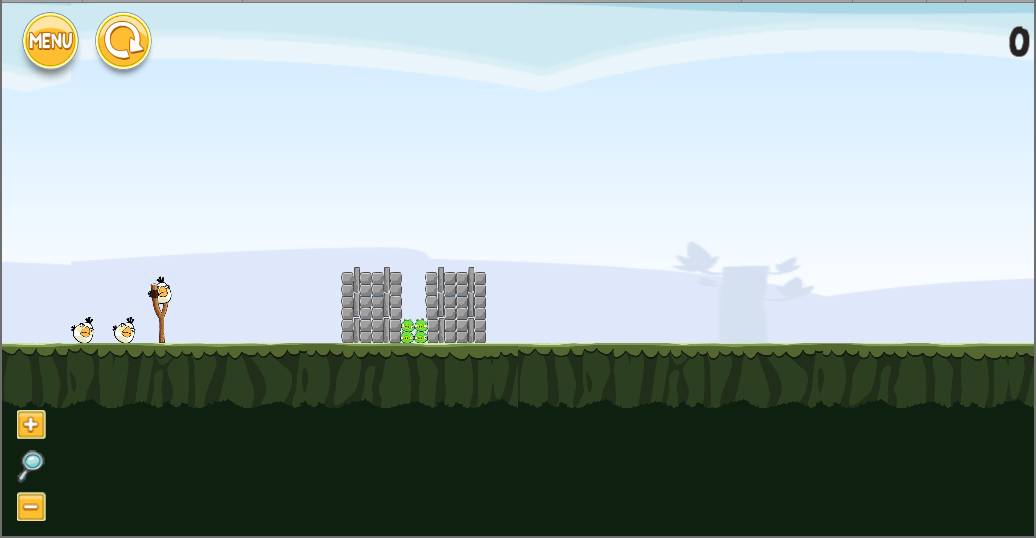}
    \caption{4}
  \end{subfigure}
  \begin{subfigure}[b]{0.32\columnwidth}
    \includegraphics[width=\linewidth]{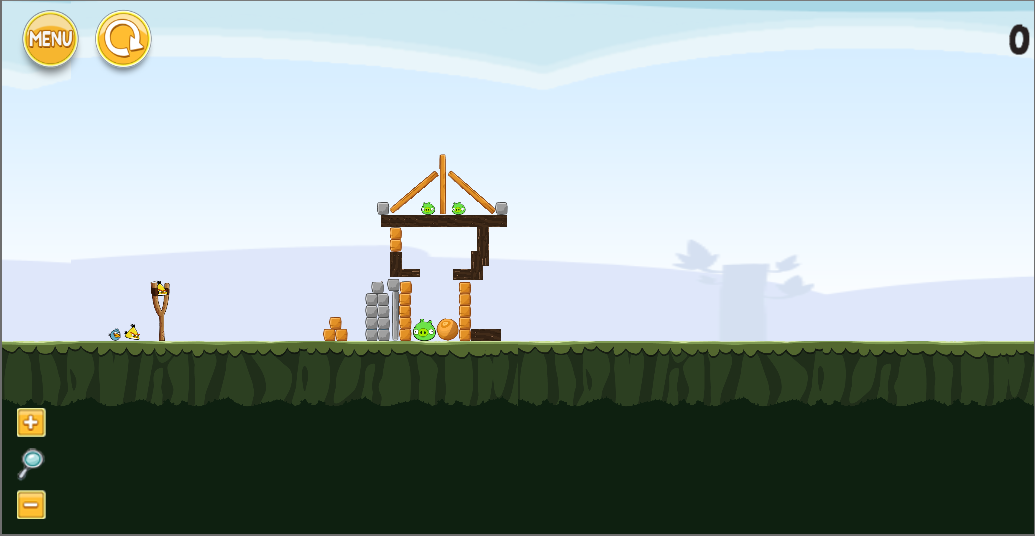}
    \caption{5}
  \end{subfigure}
  \begin{subfigure}[b]{0.32\columnwidth}
    \includegraphics[width=\linewidth]{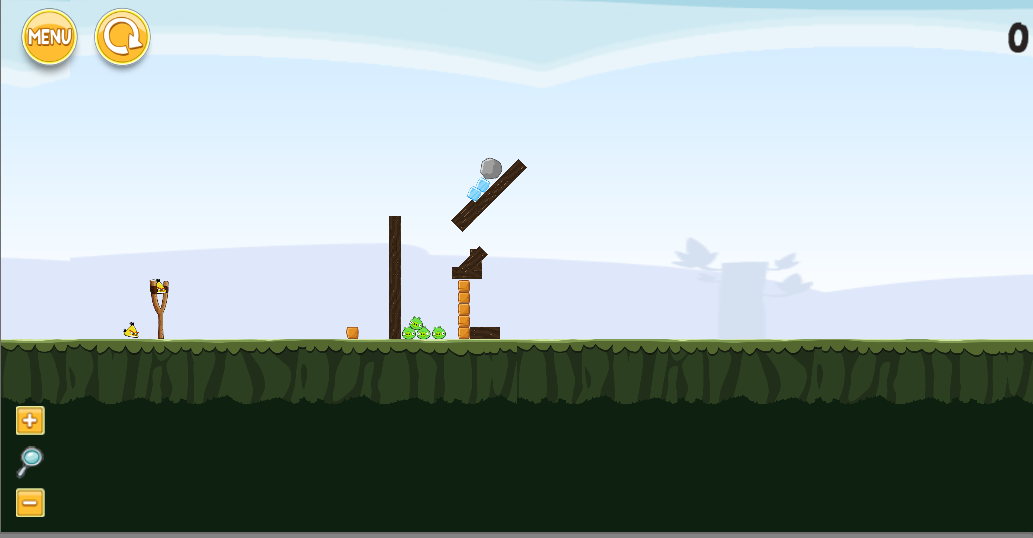}
    \caption{6}
  \end{subfigure}
  \begin{subfigure}[b]{0.32\columnwidth}
    \includegraphics[width=\linewidth]{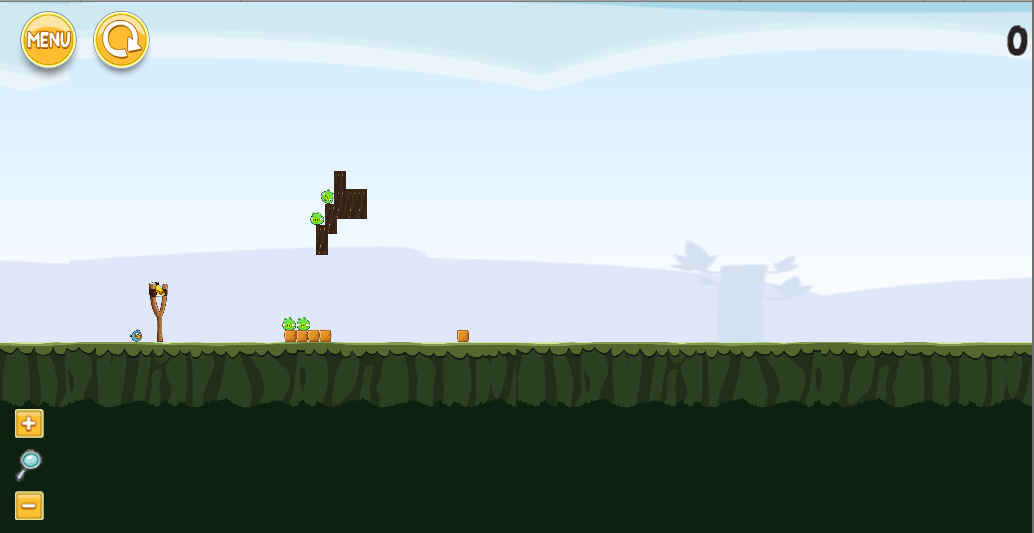}
    \caption{7}
  \end{subfigure}
  \begin{subfigure}[b]{0.32\columnwidth}
    \includegraphics[width=\linewidth]{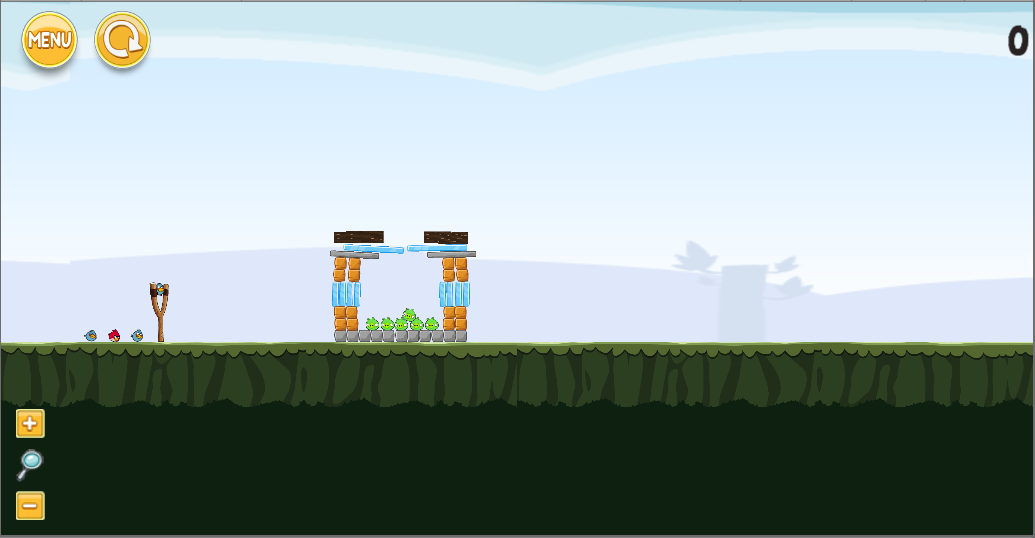}
    \caption{8}
  \end{subfigure}
\caption{The game levels from the 2021 AIBIRDS competition used for the AIBIRDS heuristic agent evaluation. The game levels were selected by excluding the ones with cartoon effects.}
\label{2021_comp_levels}
\end{figure}

\section{Relation to learning in video games}
There has been a remarkable application of AI methods [2, 3, 4, 5] on a range of video games,  from simple 2-D Atari games to more complex 3-D games such as Doom, Minecraft, StarCraft, and Dota2. It is noticeable that some of the games require a certain level of understanding of physical rules. For example, in the Atari game Breakout, an agent needs to learn to place paddle places that the ball can be rebounded and destroy as many blocks as possible. In contrast to prior work on video games, in Phy-Q we require agents to generalise to tasks that can be solved by using the same physical rule learned from the training set. 

\section{Learning Agents Implementation}
\label{appendix:agent_implementation}
\begin{itemize}
    \item \textbf{Deep Q-network (DQN): } The DQN [6] agent collects state-action-reward-next state quadruplets at the training time following decaying epsilon greedy. The quadruplets are then stored in a \textit{replay buffer} $E=\{e_1,...,e_n\}$, where each $e_i = (s_i, a_i, r_{i+1}, s_{i+1})$ is an \textit{experience}. 
    We define the reward function as task pass status, meaning the agent receives $1$ if the task is passed and $0$ otherwise. The agent uses a discretized action space of 180 actions, where each corresponds to a release degree from the slingshot with a maximum stretch. At the end of each update step, the agent trains a Deep Q-network on the sampled experiences to predict the Q-value of actions for a given state. In our experiments, we use Double Dueling Deep Q-network [7, 8], which comprises of a state encoder that transforms an input state into a hidden representation through convolutional filters, which is then separated into a state-value and advantage streams. 
    We tested four improvements upon a vanilla DQN: Double DQN [7], Dueling DQN [8], Prioritized Experience Replay [9] and Double Dueling DQN, each with two different input types: symbolic representation and screenshot. For symbolic representation, we map each game state to a $h \times w \times o_t$ tensor, where $o_t$ is the number of object types. In our experiments, we set $h=120$, $w=160$, and $o_t = 12$. For the state encoder, we use a $1\times1$ convolutional filter to squeeze the channels to $1$ and then use $h \times w$ filters to extract features from the representation that preserves crucial spatial information. For agents with screenshot state representation, we re-scale and normalise the image to $h=120$, $w=160$, and use ResNet-18 [10] to extract features and estimate Q-values. We trained both of the networks using an Adam optimizer [11] with a learning rate of $0.0003$. 
    We refer the Dueling Double DQNs we experimented with the two different input types as D-DDQN-Image and D-DDQN-Symbolic. We report only the performances of D-DDQN-Image and D-DDQN-Symbolic as they achieved the best performance compared to other variants. 
    
    \item \textbf{Policy Learners:} To support the development of the learning agents and to allow using existing RL libraries, our framework follows the OpenAI Gym [12] requirements. This allowed us to evaluate the two multi-processing agents from the Stable-Baselines3 [13]: A2C [14] and PPO [15]. We trained the two agents with both discrete and continuous action spaces, different training settings, and various hyper-parameters, but similar to [16], policy-gradient methods did not show good results compared to DQN and were unable to converge to the good policy in a reasonable time. Therefore, we exclude these two agents in the results section.
    
    \item \textbf{Relational Deep Q-network}: The relational agent consists of the relational module [17] that was built on top of the deep Q-network. The aim of this agent is to generalize over the presented templates/events by using the structured perception and relational reasoning. In our experiments, we wanted to test if the relational agent would be able to learn the important relations between the objects that could be generalized to other templates or events. We have tested the agent with symbolic and image input types and refer to them as Relational-symbolic and Relational-image agents respectively. For the image agent, we use ResNet-18 as the feature extractor and feed its output to the Multi-head dot product attention module (MHDPA) [17]. The output of the MHDPA is then used to compute the policy. For the symbolic agent, the setting is similar to the symbolic DQN agent except that the output of the convolutional layer is passed to the relational module.

\end{itemize}

\section{Computing Resources Used for the Experiments}

We used computing resources from Amazon AWS to run the evaluations. Four instances of vCPUs with 2.5 GHz processors and 16 GB memory were used. NVIDIA T4 GPUs with 16 GB memory were used in those instances.

\section*{References}

[1] Mnih, V., Kavukcuoglu, K., Silver, D., Graves, A., Antonoglou, I., Wierstra, D., \& Riedmiller, M. (2013). Playing atari with deep reinforcement learning. arXiv preprint arXiv:1312.5602.

[2] Lample, Guillaume \& Chaplot, Devendra. (2016). Playing FPS Games with Deep Reinforcement Learning. 

[3] Baker, B., Akkaya, I., Zhokhov, P., Huizinga, J., Tang, J., Ecoffet, A., ... \& Clune, J. (2022). Video PreTraining (VPT): Learning to Act by Watching Unlabeled Online Videos. arXiv preprint arXiv:2206.11795.

[4] Vinyals, O., Babuschkin, I., \& Czarnecki, W.M. (2019) Grandmaster level in StarCraft II using multi-agent reinforcement learning. Nature 575, 350–354. https://doi.org/10.1038/s41586-019-1724-z

[6] Volodymyr Mnih, Koray Kavukcuoglu, David Silver, Andrei A. Rusu, Joel Veness, Marc G. Bellemare, Alex Graves, Martin Riedmiller, Andreas K. Fidjeland, Georg Ostrovski, Stig Petersen, Charles Beattie, Amir Sadik, Ioannis Antonoglou, Helen King, Dharshan Kumaran, Daan Wierstra, Shane Legg, and Demis Hassabis. Human-level control through deep reinforcement learning. Nature, 518(7540):529–533, February 2015.

[7] Hado van Hasselt, Arthur Guez, and David Silver. Deep reinforcement learning with double q-learning. CoRR, abs/1509.06461, 2015.

[8] Ziyu Wang, Nando de Freitas, and Marc Lanctot. Dueling network architectures for deep reinforcement learning. CoRR, abs/1511.06581, 2015.

[9] Tom Schaul, John Quan, Ioannis Antonoglou, and David Silver. Prioritized experience replay. arXiv preprint arXiv:1511.05952, 2015.

[10] Kaiming He, Xiangyu Zhang, Shaoqing Ren, and Jian Sun. Deep residual learning for image recognition. CoRR, abs/1512.03385, 2015.

[11] Diederik P. Kingma and Jimmy Ba. Adam: A method for stochastic optimization. CoRR, abs/1412.6980, 2014.

[12] Greg Brockman, Vicki Cheung, Ludwig Pettersson, Jonas Schneider, J. Schulman, Jie Tang, and Wojciech Zaremba. Openai gym. ArXiv, abs/1606.01540, 2016.

[13] Antonin Raffin, Ashley Hill, Maximilian Ernestus, Adam Gleave, Anssi Kanervisto, and Noah Dormann. Stable baselines3. https://github.com/DLR-RM/stable-baselines3, 2019.

[14] V. Mnih, Adrià Puigdomènech Badia, Mehdi Mirza, A. Graves, T. Lillicrap, Tim Harley, D. Silver, and K. Kavukcuoglu. Asynchronous methods for deep reinforcement learning. In ICML, 2016.

[15] J. Schulman, F. Wolski, Prafulla Dhariwal, Alec Radford, and Oleg Klimov. Proximal policy optimization algorithms. ArXiv, abs/1707.06347, 2017.

[16] Anton Bakhtin, Laurens van der Maaten, Justin Johnson, Laura Gustafson, and Ross Girshick. Phyre: A new benchmark for physical reasoning. In NeurIPS, 2019.

[17] Vinícius Flores Zambaldi, David Raposo, Adam Santoro, Victor Bapst, Yujia Li, Igor Babuschkin, Karl Tuyls, David P. Reichert, Timothy P. Lillicrap, Edward Lockhart, Murray Shanahan, Victoria Langston, Razvan Pascanu, Matthew M. Botvinick, Oriol Vinyals, and Peter W. Battaglia. Relational deep reinforcement learning. ArXiv, abs/1806.01830, 2018.

\end{document}